\documentclass[letterpaper]{article} 
\usepackage[draft]{aaai25}  
\usepackage{times}  
\usepackage{helvet}  
\usepackage{courier}  
\usepackage[hyphens]{url}  
\usepackage{graphicx} 
\usepackage{natbib}  
\usepackage{caption} 
\usepackage{algorithm}
\usepackage{algorithmic}

\usepackage{newfloat}
\usepackage{listings}
\DeclareCaptionStyle{ruled}{labelfont=normalfont,labelsep=colon,strut=off} 
\floatstyle{ruled}
\newfloat{listing}{tb}{lst}{}
\floatname{listing}{Listing}
\usepackage{array}
\usepackage{booktabs}
\usepackage{bm}
\usepackage{caption}
\usepackage{subcaption}
\usepackage{graphicx}
\usepackage{amsmath}
\usepackage{amssymb}
\usepackage{algorithmic}
\usepackage{algorithm}
\usepackage{multirow}
\usepackage{pifont}
\usepackage{arydshln}
\usepackage{makecell}
\usepackage{xspace}
\usepackage[table]{xcolor}
\usepackage{comment}
\usepackage{rotating}

\DeclareMathOperator*{\argmax}{arg\,max}

\newcommand{\fig}{Figure\xspace}
\newcommand{\tab}{Table\xspace}
\newcommand{\algo}{Algorithm\xspace}
\newcommand{\ours}{TAFAS\xspace}
\newcommand{\tsf}{TSF\xspace}
\newcommand{\tta}{TTA\xspace}
\newcommand{\tsftta}{TSF-TTA\xspace}
\newcommand{\revin}{RevIN\xspace}
\newcommand{\dishts}{Dish-TS\xspace}
\newcommand{\san}{SAN\xspace}

\newcommand{\lb}{look-back\xspace}
\newcommand{\nvar}{C\xspace}
\newcommand{\var}{c\xspace}

\newcommand{\paas}{PAAS\xspace}

\newcommand{\equ}{Eq.}
\newcommand{\suppl}{Appendix\xspace}

\newcommand{\ie}{\textit{i.e.,~}}
\newcommand{\eg}{\textit{e.g.,~}}
\newcommand{\cmark}{\ding{51}}
\newcommand{\xmark}{\ding{55}}

\newcommand{\fmodel}{$\mathcal{F}_{\bm{\theta}}$}  

\title{Battling the Non-stationarity in Time Series Forecasting via Test-time Adaptation}
\author{
    HyunGi Kim\textsuperscript{\rm 1}, 
    Siwon Kim\textsuperscript{\rm 1}, 
    Jisoo Mok\textsuperscript{\rm 1}, 
    Sungroh Yoon\textsuperscript{\rm 1, \rm 2, \rm 3 \dag}
}
\affiliations{
    \textsuperscript{\rm 1}Department of Electrical and Computer Engineering, Seoul National University\\
    \textsuperscript{\rm 2}Interdisciplinary Program in Artificial Intelligence, Seoul National University\\
    \textsuperscript{\rm 3}AIIS, ASRI, and INMC, Seoul National University\\
    $^\dag$ Corresponding Author\\

    rlagusrl0128@snu.ac.kr, tuslkk17@gmail.com, magicshop1118@snu.ac.kr, sryoon@snu.ac.kr
}

\begin{document}

\maketitle

\begin{abstract}
Deep Neural Networks have spearheaded remarkable advancements in time series forecasting (\tsf), one of the major tasks in time series modeling. 
Nonetheless, the non-stationarity of time series undermines the reliability of pre-trained source time series forecasters in mission-critical deployment settings.
In this study, we introduce a pioneering test-time adaptation framework tailored for \tsf (\tsf-\tta). \ours, the proposed approach to \tsftta, flexibly adapts source forecasters to continuously shifting test distributions while preserving the core semantic information learned during pre-training. 
The novel utilization of partially-observed ground truth and gated calibration module enables proactive, robust, and model-agnostic adaptation of source forecasters.
Experiments on diverse benchmark datasets and cutting-edge architectures demonstrate the efficacy and generality of \ours, especially in long-term forecasting scenarios that suffer from significant distribution shifts. The code is available at \url{https://github.com/kimanki/TAFAS}.
\end{abstract}

%

\section{Introduction}
Time series forecasting (TSF), which is one of the most core tasks in time series modeling, aims to predict future values based on historical data points. 
The widespread applications of TSF across various industries include but are not limited to: weather prediction~\cite{verma2024climode}, traffic forecasting~\cite{liu2023spatio}, stock market prediction~\cite{li2023clustering}, and supply chain management~\cite{hosseinnia2023applications}.
Such a broad and over-arching impact of TSF results highlights the importance of developing a dependable time series forecaster, whose predictions maintain reliability despite changes in external factors.

\begin{figure}[t]
    \centering
        \includegraphics[width=\linewidth]{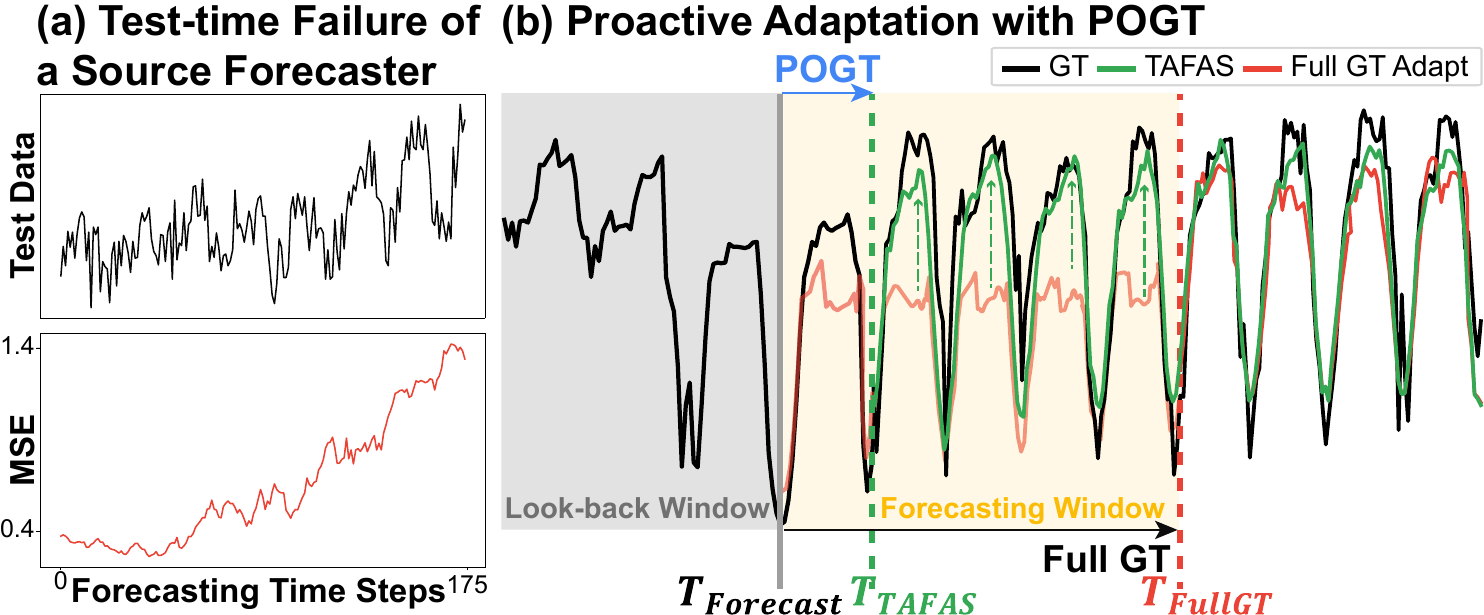}
        \caption{(a) The performance of a pre-trained source forecaster degrades when it encounters distribution-shifted inputs at test-time. In the figure, the distribution shift occurs through the gradual increase of mean value. (b) The sequential nature of time series provides the opportunity to proactively adapt the forecaster with partially-observed ground truth (POGT) before acquiring full ground truth (GT).}
    \label{fig:intro}
    \vspace{-1em}
\end{figure}

A critical bottleneck in the reliable deployment of pre-trained time series forecasters is created by the non-stationary nature of real-world time series data that leads to continuous data distribution shifts~\cite{petropoulos2022forecasting}.
Previous works on alleviating the effect of non-stationarity aim to improve the robustness of time series forecasters through advancements in the pre-training process~\cite{RevIN, Dish-TS, SAN}. 
Unfortunately, as the non-stationarity worsens the distributional discrepancy between training and test data over time, the pre-trained forecaster becomes increasingly unreliable, even if it has learned meaningful temporal semantics from training data~\cite{kuznetsov2014generalization}.
In~\fig~\ref{fig:intro}(a), we visualize how constantly evolving, non-stationary test data negatively affect the forecasting results of a pre-trained time series forecaster. 

This shortcoming of existing approaches underscores the need to continuously adapt the pre-trained source forecaster on shifted test-time inputs while preserving its core semantics.
Adapting the source forecaster to incorporate new time-variant semantics within test-time inputs allows it to reflect the ever-changing test distributions.
In this regard, we pioneer a test-time adaptation (TTA) framework tailored for TSF (\tsftta). 
\tta, which has been primarily studied in the computer vision domain under classification settings~\cite{TENT, EATA, SAR, DEYO}, dynamically adjusts a pre-trained classifier on test inputs; this objective of \tta makes it well-aligned with the aforementioned motivation of adapting the source forecaster to newly arriving test data.
Traditionally, \tta has operated under two main assumptions on the nature of test inputs. 
First, \tta assumes a complete absence of test labels because it is infeasible to hand annotate inputs at test-time.
Second, because most of the image data are assumed to be Independent and Identically Distributed (IID), \tta generally operates under the same IID assumption. 
In \tsftta, however, these assumptions no longer hold due to the intrinsic characteristics of time series data. 

Unlike the first assumption of \tta, in \tsftta, ground truth for predicted time steps eventually becomes accessible, albeit in a delayed manner. 
For instance, when predicting electricity consumption for the next 30 days, the actual amount of electricity consumption becomes known to us after 30 days. 
Interestingly, as depicted in~\fig~\ref{fig:intro}(b), the sequential nature of time series makes ground truth values partially observable before acquiring the full ground truth. In the aforementioned example, the partial ground truth for the first 7 days is obtainable only after a week. 
Utilizing this partially-observable ground truth offers an invaluable opportunity to preemptively perform \tsftta prior to the arrival of full ground truth.
Moreover, the second assumption is violated in \tsftta because temporal dependency exists in time series.
This necessitates a technique for addressing the non-IIDness of time series on local (within window) and global (throughout the entire test-time) levels.

By considering these challenges and opportunities presented by properties of time series, we propose a \textbf{T}est-time \textbf{A}daptive \textbf{F}orec\textbf{A}sting for non-stationary time \textbf{S}eries (\textbf{\ours}) that is extensible to various \tsf architectures.
TAFAS consists of periodicity-aware adaptation scheduling (\paas) and a gated calibration module (GCM).
\paas adaptively obtains partially-observed ground truth of sufficient length to represent semantically meaningful periodic patterns.
After then, model-agnostic GCMs are adapted to calibrate test-time inputs such that they conform to the distribution the source forecaster effectively handles.
The gating mechanism in GCMs controls how much the calibrated results should be utilized by considering global distribution shifts.
Together, \paas and GCM allow the source forecaster to be proactively adapted on non-stationary test-time inputs. 
Throughout the adaptation, the source forecaster remains frozen to preserve the core semantics it has learned from the extensive historical data. 
With the proactively adapted forecaster, \ours adjusts the latter part of the original predictions, where ground truths are yet to be observed, with the adapted predictions reflecting the distribution shift.

Comprehensive experimental results demonstrate that the \ours consistently enhances forecasting capabilities across source forecasters of various architectures.
\ours leads to particularly large performance gains in long-term forecasting scenarios where distribution shifts become more pronounced.
Moreover, the seamless integration of~\ours with methods addressing the non-stationarity in pre-training stage and time series foundation models further enhances their ability to navigate test-time distribution shifts. 
Notably, \ours improves the forecasting error of Chronos~\cite{ansari2024chronos} on unseen test data streams by up to 45\%.

Our contributions are summarized as follows:
\begin{itemize}
    \item We pioneer test-time adaptation in time series forecasting (\tsftta) to address the non-stationarity in time series. Our examination of the properties of time series reveals the challenges in extending existing \tta frameworks to \tsf, necessitating a new avenue to enable \tsftta.
    
    \item We introduce \ours, a model-agnostic \tsftta framework that consists of periodicity-aware adaptation scheduling (\paas) and Gated Calibration Module (GCM). These two technical components collectively enable proactive adaptation of the source forecaster on test-time inputs while preserving its core semantics.

    \item \ours consistently excels in test-time adaptation across various TSF benchmark datasets and architectures, significantly improving test errors on highly non-stationary data and in long-term forecasting scenarios.
\end{itemize}

\section{Related Works}
\label{non-stationary}
As \tsf has become a pivotal application in various industries, diverse \tsf architectures have been developed. 
Due to the page limit, an exhaustive discussion on \tsf architectures and TTA is included in \suppl. 
Here, we focus on studies that improve the time series forecaster by mitigating distribution shifts caused by the non-stationarity of time series. A line of studies introduces normalization and de-normalization modules before and after the forecaster to remove and restore the non-stationary statistics~\cite{RevIN, Stationary, Dish-TS, SAN}. 
RevIN~\cite{RevIN} performs instance normalization with learnable scale and bias factors, whereas NST~\cite{Stationary} uses a non-parametric approach without learnable transformations. 
Dish-TS~\cite{Dish-TS} and SAN~\cite{SAN} perform statistical prediction both within the look-back window and between the look-back and prediction windows to appropriately execute normalization and denormalization. 
However, their generalization capability to the continuously evolving test data distribution is inherently limited as they address non-stationarity only within training distributions. 
Although~\tsftta and online \tsf share a common ground in learning from streaming inputs~\cite{OneNet, ao2023continual, guo2016robust, FastSlow}, their primary objectives are fundamentally different.
Online \tsf aims to train a time series forecaster from scratch with streaming data, but \tsftta aims to adapt a pre-trained source forecaster. 

\section{Challenges and Opportunities in \tsf-\tta}
Generally, \tta leverages unlabeled test-time inputs to adapt classifiers on shifted distributions. 
In this study, we aim to extend \tta to \tsf to improve the ability of the pre-trained source time series forecaster to handle newly arriving test data.
While the task characteristics of \tsf and the properties of time series make applying the existing \tta frameworks to \tsf a non-trivial problem, they also open new doors to develop a tailored approach to \tsftta.
In this section, we highlight the challenges (C) and opportunities (O) unique to \tsf-\tta after introducing task definition and formulations.

\textbf{Task definition \& formulations.} 
\tsf is the task of predicting the future horizon window of $H$ time steps ($\{ \bm{x}_{t+1}, ..., \bm{x}_{t+H} \}$) given the past look-back window of $L$ time steps ($\{ \bm{x}_{t-L+1}, ..., \bm{x}_{t} \}$). $\bm{x}_t \in \mathbb{R}^\nvar$ denotes $\nvar$ number of variables observed at time $t$. To perform \tsf, a time series forecaster \fmodel $: \mathbb{R}^{L \times \nvar} \rightarrow \mathbb{R}^{H \times \nvar}$ is trained to predict subsequent future $H$ time steps given $L$ past time steps. 

The train, validation, and test sets of \tsf datasets are obtained by splitting a single continuous time series $\{ \bm{x}_1, ..., \bm{x}_T \}$ in chronological order. 
Then, the ($\texttt{past}, \texttt{future}$) pairs are obtained using a sliding window, \ie $(\bm{X}_t, \bm{Y}_t) = (\{ \bm{x}_{t-L+1}, ..., \bm{x}_{t} \}, \{ \bm{x}_{t+1}, ..., \bm{x}_{t+H} \})$. 
In non-stationary time series, data distribution continuously changes over time, resulting in distribution shifts between data splits and also within each split. 

\textbf{C1. Entropy-based \tta losses are infeasible for regression-based \tsf.} 
\tta fundamentally assumes that ground truth labels are not available. 
Therefore, existing \tta methods utilize the entropy of predicted class probability distributions to formulate objective functions for the adaptation~\cite{TENT, EATA, DEYO}. 
However, \tsf is a regression task, where the entropy of class probabilities is ill-defined, rendering the straightforward extension of entropy-based losses impossible.

\textbf{O1. Ground truth is accessible in \tsf.} In \tsf, ground truth values become accessible as the predicted future time steps eventually arrive. 
Therefore, instead of entropy-based losses, the Mean Squared Error (MSE) loss, the de facto objective function used for regression tasks, can be used as a learning signal to perform \tsftta.

\textbf{C2. Full ground truth is accessible in a delayed time, resulting in a delayed adaptation.}
However, computing the MSE loss after observing full ground truth results in a delay of $H$ time steps between the point of forecasting ($t$) and obtaining the full ground truth ($t' = t + H$).
Thus, na\"ively waiting for the arrival of full ground truth to perform \tsftta implies that none of the forecasted predictions in $H$ time steps can be adapted.
As the length of the forecasting window increases, the point at which full ground truth becomes obtainable is further delayed.
This further delay in the point of adaptation inhibits performing \tsftta in a timely manner to reflect the adjacent distribution shifts.

\textbf{O2. Utilizing partially-observed ground truth enables proactive \tsf-\tta.} The sequential nature of test-time inputs makes ground truth partially observable before acquiring the full ground truth. 
After $p$ time steps ($p<H$) from the forecasting time step $t$ of $\bm{X}_{t}$, the first $p$ time steps out of the full ground truth (\ie $\{ \bm{x}_{t+1}, \dots, \bm{x}_{t+p} \} \in \mathbb{R}^{p \times \nvar}$) are observable. 
Replacing the full ground truth in the MSE loss with its partially-observed counterpart reduces the adaptation delay and thus enables proactive adaptation.

\section{TAFAS: Test-time Adaptive Forecasting for Non-stationary Time Series}
\begin{figure*}[t]
    \centering
    \includegraphics[width=\linewidth]{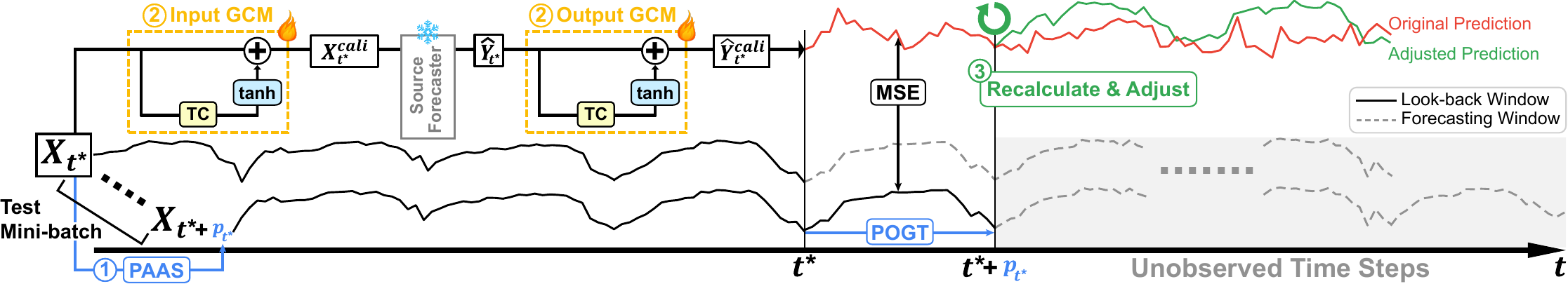}
    \vspace{-1.5em}
    \caption{An overview of \ours. \textbf{(1: Blue)} By computing the periodicity of dominant patterns for $\bm{X}_{t^*}$, \paas determines the length of partially-observed ground truth (POGT) $p_{t^*}$. \textbf{(2: Yellow)} Then input and output GCMs are proactively adapted on $\bm{X}_{t^*}$ at $t^* + p_{t^*}$ to mitigate local and global distribution shifts through Temporal Calibration (TC) and gating (tanh) mechanisms, by minimizing MSE between the POGT and corresponding prediction. The source forecaster is frozen to preserve its core semantic information learned. \textbf{(3: Green)} Following the proactive adaptation, predictions for test mini-batch $\{\bm{X}\}_{t^*}^{t^* + p_{t^*}}$ are recalculated to reflect the distribution shift and the unobserved part of the original predictions is adjusted with the adapted predictions.}
    \label{fig:overview}
\vspace{-1em}
\end{figure*}

In this section, we introduce \ours, a novel framework that considers the challenges and opportunities of~\tsftta. The overall pipeline of \ours is provided in \fig~\ref{fig:overview}.
\subsection{Periodicity-Aware Adaptation Scheduling (\paas)}
\label{paas}
To enable a proactive adaptation of the pre-trained source forecaster by reducing the adaptation delay, in~\ours, we utilize partially-observed ground truth (POGT).
However, as stated in~\textbf{Section O2.}, POGT does not eliminate the adaptation delay because to obtain POGT of length $p$, we must wait for $p$ time steps.
Therefore, choosing the appropriate value of $p$ is important for balancing the trade-off between the amount of semantic information in POGT and the adaptation delay.
When $p$ is large, the POGT contains copious semantic information, but the adaptation delay increases, offsetting the advantage of employing the POGT. 
Conversely, when $p$ is small, the forecaster can be adapted more proactively, but the POGT may contain meaningless patterns. 

To ensure that the POGT incorporates semantically meaningful temporal patterns while preventing an excessive time delay, we introduce a periodicity-aware adaptation scheduling (\paas) scheme that reflects the inherent periodic patterns of the test-time inputs. 
Several studies have demonstrated that the look-back window contains meaningful periodic patterns~\cite{Timesnet, Autoformer}.
\paas thus extracts these patterns from the look-back window to determine $p$.
From here on, $t_0$ denotes the time step at which the first test-time look-back window is obtained.
\paas applies variable-wise Fast Fourier Transform (FFT) on the first look-back window $\bm{X}_{t_0}$ to identify the variable with the highest signal power (\equ~\ref{eq:var}). 
Before FFT, the mean of $\bm{X}_{t_0}$ is set to zero to remove the influence of bias.
For the identified variable $c^*$, \paas calculates the amplitude of each frequency component to determine the dominant frequency (\equ~\ref{eq:freq}). 
\begin{align}
c^* &= \argmax_{c} \sum_{f} \lVert \mathrm{FFT}(\bm{X}_{t_0}^\var) \rVert^2 \label{eq:var} \\
f^* &= \argmax_{f} \lVert \mathrm{FFT}(\bm{X}_{t_0}^{c^*}) \rVert^2 \label{eq:freq}
\end{align}
Based on the relationship between the frequency and period, \paas derives the period of the dominant periodic patterns of $\bm{X}_{t_0}$ and set it to the length of POGT as $p_{t_0} = \Big\lceil \frac{L}{f^*} \Big\rceil$. 
The resulting periodicity-aware POGT $\bm{Y}_{t_0}[:p_{t_0}]$ includes relevant semantics embodied in the dominant periodic patterns.

Once $p_{t_0}$ is determined, $p_{t_0}+1$ instances are aggregated into a test mini-batch: $\{\bm{X}\}_{t_0}^{t_0 + p_{t_0}} = \{ \bm{X}_{t_0}, \dots, \bm{X}_{t_0+p_{t_0}} \}$. 
When the subsequent look-back window arrives at time step $t_0+p_{t_0}+1$, \paas is repeated to calculate the subsequent length of POGT adaptively.
Considering the inherent periodic patterns vary across datasets and instances, the adaptive characteristic of \paas assures data-agnostic \tsftta.

\subsection{Gated Calibration Module (GCM)}
Selecting the module to adapt is a critical design choice in \tta.
Existing \tta methods generally adapt normalization layers,~\textit{e.g.,} Batch Normalization~\cite{BatchNorm}, to adjust the distributions of the intermediate features. 
However, state-of-the-art \tsf methods adopt various forms of architectures, many of which are missing such normalization layers.
Hence, we introduce a model-agnostic Gated Calibration Module (GCM) to guarantee that~\ours can be applied generally to diverse pre-trained source forecasters.

\ours adapts the GCM with the source forecaster frozen to preserve the core temporal semantics that the forecaster has learned from extensive historical data. 
GCM is attached to both the front and tail ends of the source forecaster, referred to as input and output GCMs. 
The input GCM maps the distribution-shifted test input $\bm{X}_{t}$ to a calibrated input $\bm{X}_{t}^{\textrm{cali}}$ that belong in a distribution the source forecaster can handle. 
The output GCM remaps the source forecaster's prediction $\bm{\hat{Y}}_{t}$ to $\bm{\hat{Y}}_{t}^{\textrm{cali}}$ in order to calibrate $\bm{\hat{Y}}_{t}$ back to the continuously changing test distribution.

GCM consists of variable-wise temporal calibration and gating mechanisms to handle both the local (within the look-back window) and global (throughout the entire test-time) non-stationarity.
The temporal calibration handles local distribution shifts by transforming the given window to calculate calibrated results.
The gating mechanism handles global distribution shifts by updating how much to reflect calibrated results over time. 
Both temporal calibration and gating mechanisms are applied variable-wise, since each variable can have a different degree of non-stationarity. 
The two operations in the input GCM are expressed as the following:
\begin{equation}\label{eq:gcm}
\begin{split}
\textrm{GCM}(\bm{X}_{t}) = \bm{X}_{t} &+ \\
\mathrm{Tile}(\tanh(\bm{\alpha})) &\circ 
\Big( \mathrm{Concat}(\{\bm{W}^c \bm{X}_{t}^c\}_{c=1}^C) + \bm{b} \Big),
\end{split}
\end{equation}
where $\bm{W}^c \in \mathbb{R}^{L \times L}$, $\bm{b} \in \mathbb{R}^{L \times \nvar}$, and $\bm{\alpha} \in \mathbb{R}^{\nvar}$. $\mathrm{Tile(\cdot)}: \mathbb{R}^{\nvar} \rightarrow \mathbb{R}^{L \times \nvar}$ broadcasts the gating vector in a temporal dimension, $\mathrm{Concat}(\cdot)$ concatenates the calibrated signals along variable dimension, and $\circ$ denotes a Hadamard product. $\bm{W}^c$ and $\bm{b}$ are initialized to be zero so that at the first test time step when the test data diverge little from the training distribution, the original input is passed without calibration.
The mechanism of the output GCM follows~\equ~\ref{eq:gcm} with $\bm{X}_t$ replaced with $\hat{\bm{Y}}_t$, and the dimensions of $\bm{W}^c$ and $\bm{b}$ adjusted accordingly to $H \times H$ and $H \times C$.

Let $t^*$ denote a time step at which \paas calculates the POGT length ($t_0, t_0 + p_{t_0} + 1, \dots$) and $p_{t^*}$ denote the POGT length computed at $t^*$.
After a test mini-batch $\{\bm{X}\}_{t^*}^{t^* + p_{t^*}}$ is obtained at $t^*+p_{t^*}$, GCMs are adapted by minimizing the TAFAS loss defined as the following:
\begin{align}
\mathcal{L}^{\textrm{partial}} &= \mathrm{MSE}(\bm{\hat{Y}}_{t^*}^{\textrm{cali}}\left[ :p_{t^*} \right], \bm{Y}_{t^*} \left[ :p_{t^*} \right]) \\ 
\mathcal{L}^{\textrm{full}} &= \mathrm{MSE}(\{ \bm{\hat{Y}}^{\textrm{cali}} \}_{\tilde{t}^*}^{\tilde{t}^*+p_{\tilde{t}^*}}, \{\bm{Y} \}_{\tilde{t}^*}^{\tilde{t}^*+p_{\tilde{t}^*}}) \\
\mathcal{L}^{\textrm{\ours}} &= \mathcal{L}^{\textrm{partial}} + \mathcal{L}^{\textrm{full}},
\end{align}
where $\tilde{t}^*$ represents the most recent time step among the past POGT-computing steps whose corresponding mini-batches have now observed their full ground truths.
$\mathcal{L}^{\textrm{partial}}$ is computed between the first $p_{t^*}$ time steps of the calibrated prediction for $\bm{X}_{t^*}$ whose periodicity-aware POGT is available $(\bm{\hat{Y}}_{t^*}^{\textrm{cali}}\left[ :p_{t^*} \right])$ and the corresponding POGT $(\bm{Y}_{t^*} \left[ :p_{t^*} \right])$. 
$\mathcal{L}^{\textrm{full}}$ is calculated between the calibrated predictions of all look-back windows within the past mini-batch constructed at $\tilde{t}^*+p_{\tilde{t}^*}$ and the associated full ground truths to use their longer semantic information for adaptation.
This adaption process of \ours enables proactive adaptation right when the semantically meaningful POGT is obtained while utilizing the extended semantic details from the full ground truths of the past mini-batch. 
When no past mini-batch with full ground truths is available, \eg at the beginning of the adaptation phase,~\ours uses only $\mathcal{L}^{\textrm{partial}}$ for adaptation. 

\subsection{Prediction Adjustment (PA)}
Because the forecaster is proactively adapted,~\ours can replace the latter part of original predictions, whose ground truths are yet to be observed, with adjusted predictions that reflect the distribution shift. 
After the forecaster is adapted at $t^*+p_{t^*}$, ~\ours recalculates the predictions for all look-back windows in $\{\bm{X}\}_{t^*}^{t^* + p_{t^*}}$ and then substitutes the original predictions for time steps after $t^* + p_{t^*}$ with the adapted predictions.
Specifically, for the \lb window $\bm{X}_{t^*+k}$, where $k \in \{ 0, \dots, p_{t^*} \}$, the corresponding prediction $\hat{\bm{Y}}_{t^*+k}^{\textrm{cali}}$ predicts time steps $\{ (t^*+k+1), \dots, (t^*+k+H) \}$. For the time steps $\{ (t^* + p_{t^*}+1), \dots, (t^*+k+H) \}$ which are yet to be observed, \ours substitutes the original prediction $\hat{\bm{Y}}_{t^*+k}^{\textrm{cali}}$ with the adapted prediction $\hat{\bm{Y}}_{t^*+k}^{\textrm{cali, adapted}}$ that reflects distribution shifts as the following:
\begin{equation}
\hat{\bm{Y}}_{t^*+k,\, i}^{\textrm{adjust}} = 
\begin{cases}
    \hat{\bm{Y}}_{t^*+k,\, i}^{\textrm{cali}} & \text{if } i \leq (t^* + p_{t^*}) \\ 
    \hat{\bm{Y}}_{t^*+k,\, i}^{\textrm{cali, adapted}} & \text{if } i > (t^* + p_{t^*}).
\end{cases}
\end{equation}
$\hat{\bm{Y}}_{t^*+k,\, i}^{\textrm{cali, adapted}}$ denotes the adapted prediction values for the time step $i$ of $\hat{\bm{Y}}_{t^*+k}^{\textrm{cali}}$. 
We summarize the overall pipeline of \ours in \suppl.

\begin{table*}[ht]
    \tiny
    \renewcommand{\arraystretch}{0.8}
    \setlength{\tabcolsep}{5.5pt}
    \centering
    \caption{Test MSE on multivariate time series forecasting datasets with and without \ours across various \tsf architectures: Transformer-based (iTransformer, PatchTST), Linear-based (DLinear, OLS), and MLP-based (FreTS, MICN). $+$\ours denotes whether the \ours framework is applied to the corresponding source forecaster. The lower MSE is marked in bold.}
    \vspace{-0.5em}
    \resizebox{\textwidth}{!}{
    \begin{tabular}{cc|cccc|cccc|cccc}
        \toprule
        \multicolumn{2}{c}{} & \multicolumn{4}{c}{Transformer-based} & \multicolumn{4}{c}{Linear-based} & \multicolumn{4}{c}{MLP-based} \\
        \cmidrule(lr){3-6} \cmidrule(lr){7-10} \cmidrule(lr){11-14}
        \multicolumn{2}{c}{Models} & \multicolumn{2}{c}{iTransformer} & \multicolumn{2}{c}{PatchTST} & \multicolumn{2}{c}{DLinear} & \multicolumn{2}{c}{OLS} & \multicolumn{2}{c}{FreTS} & \multicolumn{2}{c}{MICN} \\
        \multicolumn{2}{c}{$+$ \ours} & \xmark & \cmark & \xmark & \cmark & \xmark & \cmark & \xmark & \cmark & \xmark & \cmark & \xmark & \cmark \\
        \bottomrule
        \toprule
        \multirow{4}{*}{\rotatebox[origin=c]{90}{ETTh1}} & 96 & 0.444 & \textbf{0.438} & 0.436 & \textbf{0.429} & 0.451 & \textbf{0.442} & 0.451 & \textbf{0.441} & 0.441 & \textbf{0.437} & 0.455 & \textbf{0.446} \\
         & 192 & 0.503 & \textbf{0.492} & 0.492 & \textbf{0.481} & 0.504 & \textbf{0.493} & 0.504 & \textbf{0.494} & 0.498 & \textbf{0.491} & 0.513 & \textbf{0.501} \\
         & 336 &  0.562 & \textbf{0.554} & 0.539 & \textbf{0.529} & 0.551 & \textbf{0.541} & 0.551 & \textbf{0.540} & 0.563 & \textbf{0.555} & 0.574 & \textbf{0.561} \\
         & 720 & 0.786 & \textbf{0.704} & 0.713 & \textbf{0.690} & 0.700 & \textbf{0.669} & 0.700 & \textbf{0.662} & 0.715 & \textbf{0.684} & 0.736 & \textbf{0.702} \\
        \midrule
        \multirow{4}{*}{\rotatebox[origin=c]{90}{ETTm1}} & 96 & 0.388 & \textbf{0.362} & 0.386 & \textbf{0.377} & 0.371 & \textbf{0.351} & 0.371 & \textbf{0.353} & 0.367 & \textbf{0.355} & 0.398 & \textbf{0.372} \\
         & 192 & 0.448 & \textbf{0.429} & 0.440 & \textbf{0.429} & 0.443 & \textbf{0.418} & 0.444 & \textbf{0.417} & 0.429 & \textbf{0.418} & 0.448 & \textbf{0.428} \\
         & 336 & 0.519 & \textbf{0.493} & 0.500 & \textbf{0.487} & 0.518 & \textbf{0.481} & 0.518 & \textbf{0.479} & 0.493 & \textbf{0.476} & 0.524 & \textbf{0.493} \\
         & 720 & 0.592 & \textbf{0.560} & 0.562 & \textbf{0.542} & 0.592 & \textbf{0.549} & 0.592 & \textbf{0.549} & 0.560 & \textbf{0.539} & 0.602 & \textbf{0.567} \\
        \midrule
        \multirow{4}{*}{\rotatebox[origin=c]{90}{ETTh2}} & 96 & 0.241 & \textbf{0.239} & 0.233 & \textbf{0.232} & 0.229 & \textbf{0.227} & 0.231 & \textbf{0.229} & 0.234 & \textbf{0.232} & 0.234 & \textbf{0.231} \\
         & 192 & 0.291 & \textbf{0.287} & 0.282 & \textbf{0.277} & 0.283 & \textbf{0.281} & 0.284 & \textbf{0.281} & 0.286 & \textbf{0.282} & 0.285 & \textbf{0.282} \\
         & 336 & 0.333 & \textbf{0.326} & 0.328 & \textbf{0.318} & 0.325 & \textbf{0.317} & 0.326 & \textbf{0.317} & 0.328 & \textbf{0.318} & 0.330 & \textbf{0.320} \\
         & 720 & 0.415 & \textbf{0.393} & 0.416 & \textbf{0.396} & 0.415 & \textbf{0.391} & 0.416 & \textbf{0.386} & 0.420 & \textbf{0.390} & 0.414 & \textbf{0.398} \\
        \midrule
        \multirow{4}{*}{\rotatebox[origin=c]{90}{ETTm2}} & 96 & 0.159 & \textbf{0.157} & 0.157 & \textbf{0.156} & 0.159 & \textbf{0.158} & 0.160 & \textbf{0.159} & 0.158 & \textbf{0.156} & 0.161 & \textbf{0.160} \\
         & 192 & 0.197 & \textbf{0.192} & \textbf{0.194} & \textbf{0.194} & 0.193 & \textbf{0.191} & 0.194 & \textbf{0.192} & 0.193 & \textbf{0.192} & 0.195 & \textbf{0.193} \\
         & 336 & 0.244 & \textbf{0.235} & 0.234 & \textbf{0.232} & 0.232 & \textbf{0.229} & 0.233 & \textbf{0.230} & 0.233 & \textbf{0.230} & 0.235 & \textbf{0.232} \\
         & 720 & 0.312 & \textbf{0.301} & 0.307 & \textbf{0.299} & 0.306 & \textbf{0.297} & 0.307 & \textbf{0.298} & 0.302 & \textbf{0.293} & 0.308 & \textbf{0.299} \\
         \midrule
        \multirow{4}{*}{\rotatebox[origin=c]{90}{Exchange}} & 96 & 0.086 & \textbf{0.084} & 0.084 & \textbf{0.081} & \textbf{0.078} & 0.079 & 0.081 & \textbf{0.079} & 0.084 & \textbf{0.079} & 0.083 & \textbf{0.079} \\
         & 192 & 0.175 & \textbf{0.165} & 0.179 & \textbf{0.168} & 0.171 & \textbf{0.161} & 0.172 & \textbf{0.162} & 0.176 & \textbf{0.163} & 0.183 & \textbf{0.170} \\
         & 336 & 0.329 & \textbf{0.280} & 0.350 & \textbf{0.286} & 0.321 & \textbf{0.280} & 0.323 & \textbf{0.265} & 0.326 & \textbf{0.283} & 0.342 & \textbf{0.296} \\
         & 720 & 0.844 & \textbf{0.773} & 0.845 & \textbf{0.844} & 0.837 & \textbf{0.711} & 0.836 & \textbf{0.583} & 0.840 & \textbf{0.772} & 1.276 & \textbf{0.942} \\
        \midrule
        \multirow{4}{*}{\rotatebox[origin=c]{90}{Illness}} & 24     & \textbf{2.119} & 2.124 & \textbf{2.078} & \textbf{2.078}       & 2.642 & \textbf{2.631} & 2.509 & \textbf{2.506}       & \textbf{2.516} & \textbf{2.516} & \textbf{3.280} & 3.306 \\
         & 36 & 1.989 & \textbf{1.988} & \textbf{2.095} & \textbf{2.095} & 2.501 & \textbf{2.450} & 2.435 & \textbf{2.384} & \textbf{2.441} & \textbf{2.441} & \textbf{3.503} & 3.524 \\
         & 48 & 2.173 & \textbf{2.155} & \textbf{1.964} & \textbf{1.964} & 2.487 & \textbf{2.403} & 2.417 & \textbf{2.336} & 2.257 & \textbf{2.198} & 2.066 & \textbf{2.033} \\
         & 60 & 1.925 & \textbf{1.876} & \textbf{1.833} & \textbf{1.833} & 2.530 & \textbf{2.444} & 2.490 & \textbf{2.415} & 2.074 & \textbf{2.056} & 1.915 & \textbf{1.881} \\
        \midrule
        \multirow{4}{*}{\rotatebox[origin=c]{90}{Weather}} & 96 & 0.179 & \textbf{0.170} & 0.174 & \textbf{0.171} & 0.195 & \textbf{0.179} & 0.196 & \textbf{0.179} & 0.181 & \textbf{0.169} & 0.176 & \textbf{0.175} \\
         & 192 & 0.227 & \textbf{0.214} & 0.221 & \textbf{0.215} & 0.240 & \textbf{0.223} & 0.240 & \textbf{0.223} & 0.225 & \textbf{0.212} & 0.224 & \textbf{0.222} \\
         & 336 & 0.284 & \textbf{0.265} & 0.276 & \textbf{0.265} & 0.292 & \textbf{0.270} & 0.292 & \textbf{0.270} & 0.280 & \textbf{0.260} & 0.280 & \textbf{0.274} \\
         & 720 & 0.360 & \textbf{0.343} & 0.352 & \textbf{0.334} & 0.364 & \textbf{0.345} & 0.364 & \textbf{0.343} & 0.356 & \textbf{0.336} & 0.350 & \textbf{0.349} \\
        \bottomrule
        \label{tab:main}
    \end{tabular}
    }
\vspace{-2em}
\end{table*}

\section{Experiments}
\subsection{Experimental Setup}
\label{setup}
\textbf{Datasets.} We demonstrate the effectiveness of \ours using the seven widely used multivariate \tsf benchmark datasets: ETTh1, ETTm1, ETTh2, ETTm2, Exchange, Illness, and Weather~\cite{Autoformer}. In~\suppl, we report the results of the ADF test~\cite{elliott1992efficient} to demonstrate that a substantial degree of non-stationarity exists in all these datasets.\\
\textbf{Time series forecasters.} For the source time series forecasters, we adopt six state-of-the-art forecasters across various architectures: Transformer-based (iTransformer~\cite{iTransformer}, PatchTST~\cite{PatchTST}), Linear-based (DLinear~\cite{DLinear}, OLS~\cite{OLS}), and MLP-based (FreTS~\cite{FreTS}, MICN~\cite{MICN}). 
Moreover, we verify the effectiveness of \ours on \tsf foundation model~\cite{ansari2024chronos} pre-trained on a large corpus of time series data. \ours can be applied in combination with all of these forecasters regardless of the architecture design due to its fully model-agnostic design.\\
\textbf{Implementation details.} Unless stated otherwise, we follow the standard protocol in \tsf evaluation~\cite{Timesnet}. We use the look-back window length $L = 36$ for Illness and $L = 96$ for the other datasets. For forecasting window length $H$, we evaluate on 4 different lengths, $H \in \{ 24, 36, 48, 60 \}$ for Illness and $H \in \{ 96, 192, 336, 720 \}$ for the other datasets. We split datasets in chronological order with the ratio of (0.6, 0.2, 0.2) for ETTh1, ETTm1, ETTh2, and ETTm2 and (0.7, 0.1, 0.2) for Exchange, Illness, and Weather to construct train, validation, and test sets. We repeat each pre-training run over three different seeds and select the pre-trained source forecaster with the lowest average validation MSE. More details on training processes are provided in \suppl.
\subsection{\ours on Various \tsf Architectures}

\begin{table*}[t]
    \small
    \renewcommand{\arraystretch}{0.9}
    \setlength{\tabcolsep}{2.5pt}
    \centering
    \caption{Compatability of \ours with various normalization modules addressing non-stationarity in the pre-training stage. We report test MSE with and without \ours where each source forecaster (iTransformer, DLinear, and FreTS) is pre-trained with the normalization modules of RevIN, Dish-TS, or SAN.}
    \vspace{-0.5em}

    \begin{tabular}{cc|cc|cc|cc|cc|cc|cc|cc|cc|cc}
        \toprule
        \multicolumn{2}{c|}{Models} & \multicolumn{6}{c}{iTransformer} & \multicolumn{6}{c}{DLinear} & \multicolumn{6}{c}{FreTS} \\
        \cmidrule(lr){3-8} \cmidrule(lr){9-14} \cmidrule(lr){15-20}
        \multicolumn{2}{c|}{Norm.} & \multicolumn{2}{c}{\revin} & \multicolumn{2}{c}{\dishts} & \multicolumn{2}{c}{\san} & \multicolumn{2}{c}{\revin} & \multicolumn{2}{c}{\dishts} & \multicolumn{2}{c}{\san} & \multicolumn{2}{c}{\revin} & \multicolumn{2}{c}{\dishts} & \multicolumn{2}{c}{\san} \\
        \multicolumn{2}{c|}{$+$ \ours} & \xmark & \cmark & \xmark & \cmark & \xmark & \cmark & \xmark & \cmark & \xmark & \cmark & \xmark & \cmark & \xmark & \cmark & \xmark & \cmark & \xmark & \cmark \\ 
        \bottomrule
        \toprule
        \multirow{4}{*}{\rotatebox[origin=c]{90}{ETTh1}} & 96 & 0.444 & \textbf{0.437} & 0.457 & \textbf{0.441} & 0.451 & \textbf{0.439} & 0.451 & \textbf{0.444} & 0.452 & \textbf{0.439} & 0.438 & \textbf{0.433} & 0.448 & \textbf{0.436} & 0.440 & \textbf{0.421} & 0.438 & \textbf{0.434} \\
         & 192 & 0.503 & \textbf{0.490} & 0.507 & \textbf{0.499} & 0.511 & \textbf{0.497} & 0.504 & \textbf{0.497} & 0.507 & \textbf{0.488} & 0.489 & \textbf{0.480} & 0.511 & \textbf{0.491} & 0.494 & \textbf{0.472} & 0.485 & \textbf{0.478} \\
         & 336 & 0.561 & \textbf{0.548} & 0.560 & \textbf{0.549} & 0.596 & \textbf{0.564} & 0.550 & \textbf{0.548} & 0.555 & \textbf{0.531} & 0.534 & \textbf{0.527} & 0.567 & \textbf{0.551} & 0.547 & \textbf{0.521} & 0.529 & \textbf{0.523} \\
         & 720 & 0.785 & \textbf{0.702} & 0.720 & \textbf{0.669} & 0.727 & \textbf{0.689} & 0.699 & \textbf{0.669} & 0.708 & \textbf{0.642} & 0.664 & \textbf{0.652} & 0.729 & \textbf{0.682} & 0.706 & \textbf{0.641} & 0.665 & \textbf{0.640} \\
        \midrule
        \multirow{4}{*}{\rotatebox[origin=c]{90}{ETTm1}} & 96  & 0.754 & \textbf{0.652} & 0.418 & \textbf{0.367} & 0.360 & \textbf{0.351} & 0.371 & \textbf{0.354} & 0.371 & \textbf{0.350} & 0.353 & \textbf{0.345} &  1.071 & \textbf{0.360} & 0.375 & \textbf{0.354} & 0.355 & \textbf{0.347}\\
         & 192 & 0.817 & \textbf{0.791} & 0.487 & \textbf{0.430} & 0.418 & \textbf{0.408} & 0.445 & \textbf{0.422} & 0.445 & \textbf{0.416} & 0.415 & \textbf{0.405} & 0.893 & \textbf{0.421} & 0.447 & \textbf{0.418} & 0.415 & \textbf{0.405}\\
         & 336 & 0.870 & \textbf{0.833} & 0.543 & \textbf{0.493} & 0.480 & \textbf{0.463} & 0.520 & \textbf{0.484} & 0.520 & \textbf{0.477} & 0.474 & \textbf{0.460} & 0.556 & \textbf{0.478} & 0.522 & \textbf{0.478} & 0.472 & \textbf{0.460}\\
         & 720 & 0.940 & \textbf{0.897} & 0.606 & \textbf{0.554} & 0.530 & \textbf{0.517} & 0.593 & \textbf{0.553} & 0.597 & \textbf{0.539} & 0.526 & \textbf{0.514} & 0.599 & \textbf{0.535} & 0.598 & \textbf{0.541} & 0.525 & \textbf{0.512}\\
        \midrule
        \multirow{4}{*}{\rotatebox[origin=c]{90}{ETTh2}} & 96  & 0.241 & \textbf{0.240} & 0.263 & \textbf{0.259} & 0.243 & \textbf{0.242} & 0.230 & \textbf{0.228} & 0.232 & \textbf{0.231} & 0.228 & \textbf{0.228} & 0.244 & \textbf{0.241} & 0.247 & \textbf{0.246} & 0.239 & \textbf{0.239}\\
         & 192 & 0.295 & \textbf{0.292} & \textbf{0.308} & 0.335 & 0.303 & \textbf{0.300} & 0.284 & \textbf{0.279} & 0.286 & \textbf{0.276} & 0.277 & \textbf{0.276} & 0.290 & \textbf{0.283} & 0.304 & \textbf{0.288} & 0.282 & \textbf{0.281}\\
         & 336 & 0.333 & \textbf{0.325} & \textbf{0.354} & 0.367 & 0.336 & \textbf{0.328} & 0.325 & \textbf{0.314} & 0.328 & \textbf{0.298} & 0.308 & \textbf{0.305} & 0.333 & \textbf{0.318} & 0.336 & \textbf{0.303} & 0.307 & \textbf{0.305}\\
         & 720 & 0.415 & \textbf{0.392} & 0.469 & \textbf{0.424} & 0.434 & \textbf{0.405} & 0.409 & \textbf{0.383} & 0.424 & \textbf{0.358} & 0.381 & \textbf{0.364} & 0.422 & \textbf{0.384} & 0.435 & \textbf{0.366} & 0.368 & \textbf{0.356}\\
        \midrule
        \multirow{4}{*}{\rotatebox[origin=c]{90}{ETTm2}} & 96  & 0.179 & \textbf{0.178} & 0.165 & \textbf{0.162} & 0.156 & \textbf{0.156} & 0.160 & \textbf{0.159} & 0.160 & \textbf{0.159} & 0.161 & \textbf{0.155} & 0.173 & \textbf{0.154} & 0.159 & \textbf{0.158} & 0.161 & \textbf{0.155}\\
         & 192 & 0.204 & \textbf{0.202} & \textbf{0.196} & 0.201 & 0.190 & \textbf{0.189} & 0.193 & \textbf{0.192} & 0.195 & \textbf{0.193} & 0.197 & \textbf{0.190} & 0.210 & \textbf{0.189} & \textbf{0.193} & 0.196 & 0.201 & \textbf{0.192}\\
         & 336 & 0.245 & \textbf{0.242} & 0.258 & \textbf{0.257} & 0.228 & \textbf{0.225} & 0.232 & \textbf{0.232} & 0.234 & \textbf{0.233} & 0.237 & \textbf{0.229} & 0.247 & \textbf{0.228} & \textbf{0.235} & 0.241 & 0.238 & \textbf{0.230}\\
         & 720 & 0.316 & \textbf{0.309} & \textbf{0.315} & 0.328 & 0.299 & \textbf{0.292} & 0.306 & \textbf{0.301} & 0.306 & \textbf{0.298} & 0.296 & \textbf{0.291} & 0.306 & \textbf{0.297} & 0.312 & \textbf{0.308} & 0.297 & \textbf{0.289}\\
         \midrule
        \multirow{4}{*}{\rotatebox[origin=c]{90}{Exchange}} & 96  & 0.086 & \textbf{0.084} & \textbf{0.091} & 0.119 & 0.079 & \textbf{0.078} & 0.081 & \textbf{0.078} & 0.081 & \textbf{0.076} & 0.079 & \textbf{0.078} & 0.084 & \textbf{0.079} & 0.084 & \textbf{0.080} & 0.079 & \textbf{0.078}\\
         & 192 & 0.175 & \textbf{0.164} & \textbf{0.199} & 0.310 & 0.161 & \textbf{0.155} & 0.169 & \textbf{0.164} & 0.167 & \textbf{0.149} & 0.163 & \textbf{0.158} & 0.177 & \textbf{0.164} & 0.187 & \textbf{0.162} & 0.161 & \textbf{0.158}\\
         & 336 & 0.329 & \textbf{0.282} & \textbf{0.366} & 0.546 & 0.307 & \textbf{0.275} & 0.317 & \textbf{0.293} & 0.307 & \textbf{0.254} & 0.300 & \textbf{0.275} & 0.328 & \textbf{0.302} & 0.336 & \textbf{0.295} & 0.298 & \textbf{0.276}\\
         & 720 & 0.844 & \textbf{0.557} & \textbf{0.919} & 1.529 & 1.110 & \textbf{0.645} & 0.834 & \textbf{0.815} & 0.929 & \textbf{0.546} & 0.845 & \textbf{0.802} & 0.837 & \textbf{0.738} & 0.812 & \textbf{0.675} & 0.846 & \textbf{0.704}\\
        \midrule
        \multirow{4}{*}{\rotatebox[origin=c]{90}{Illness}} & 24  & \textbf{2.118} & 2.121 & 2.765 & \textbf{2.571} & \textbf{2.587} & 2.589 & 2.654 & \textbf{2.613} & \textbf{2.778} & 2.865 & 2.460 & \textbf{2.423} & 2.480 & \textbf{2.438} & 2.577 & \textbf{2.542} & 2.536 & \textbf{2.500}\\
         & 36 & 1.989 & \textbf{1.984} & 2.701 & \textbf{2.494} & 2.492 & \textbf{2.465} & 2.503 & \textbf{2.441} & \textbf{2.606} & 2.702 & 2.513 & \textbf{2.472} & 2.420 & \textbf{2.386} & \textbf{2.530} & 2.661 & 2.477 & \textbf{2.431}\\
         & 48 & 2.183 & \textbf{2.132} & 2.527 & \textbf{2.409} & 2.386 & \textbf{2.292} & 2.487 & \textbf{2.406} & \textbf{2.525} & 2.702 & 2.443 & \textbf{2.392} & 2.247 & \textbf{2.153} & \textbf{2.246} & 2.434 & 2.416 & \textbf{2.340}\\
         & 60 & 2.030 & \textbf{2.029} & 3.372 & \textbf{2.738} & 2.363 & \textbf{2.298} & 2.529 & \textbf{2.452} & \textbf{2.549} & 2.785 & 2.423 & \textbf{2.383} & 2.081 & \textbf{1.997} & \textbf{2.068} & 2.453 & 2.408 & \textbf{2.354}\\
        \midrule
        \multirow{4}{*}{\rotatebox[origin=c]{90}{Weather}} & 96  & 0.205 & \textbf{0.199} & 0.183 & \textbf{0.164} & 0.168 & \textbf{0.163} & 0.198 & \textbf{0.184} & 0.195 & \textbf{0.178} & 0.171 & \textbf{0.165} & 0.195 & \textbf{0.165} & 0.193 & \textbf{0.166} & 0.169 & \textbf{0.164}\\
         & 192 & 0.256 & \textbf{0.246} & 0.231 & \textbf{0.209} & 0.213 & \textbf{0.203} & 0.243 & \textbf{0.225} & 0.240 & \textbf{0.219} & 0.214 & \textbf{0.207} & 0.251 & \textbf{0.204} & 0.240 & \textbf{0.206} & 0.212 & \textbf{0.204}\\
         & 336 & 0.306 & \textbf{0.286} & 0.286 & \textbf{0.257} & 0.266 & \textbf{0.253} & 0.295 & \textbf{0.269} & 0.292 & \textbf{0.267} & 0.269 & \textbf{0.258} & 0.304 & \textbf{0.253} & 0.293 & \textbf{0.253} & 0.266 & \textbf{0.256}\\
         & 720 & 0.376 & \textbf{0.356} & 0.364 & \textbf{0.326} & 0.340 & \textbf{0.325} & 0.367 & \textbf{0.342} & 0.366 & \textbf{0.337} & 0.342 & \textbf{0.335} & 0.379 & \textbf{0.331} & 0.367 & \textbf{0.321} & 0.338 & \textbf{0.328}\\
        \bottomrule
        \label{tab:norm}
    \end{tabular}
\vspace{-2em}
\end{table*}

\tab~\ref{tab:main} presents the MSE of forecasting results with and without \ours across various source \tsf architectures and multiple forecasting windows.
The full results, including Mean Absolute Error (MAE) and standard deviations, are reported in \suppl due to the space limit.
\ours consistently reduces the forecasting error at test-time, effectively handling the test-time non-stationarity of time series. 
We highlight that the effectiveness of \ours at mitigating test-time distribution shifts remains strong across various architectures and datasets, consolidating its broad model- and data-agnostic applicability.
Furthermore, when $H = 336$, \ours improves the average MSE of iTransformer and DLinear by \textbf{4.95\%} and \textbf{5.20\%}, respectively, and on an even longer forecasting window ($H = 720$), the performance improvement brought upon by \ours reaches \textbf{5.76\%} and \textbf{6.30\%}. 
These results indicate that \ours is particularly advantageous in long-term forecasting scenarios with more severe distribution shifts. 

To further demonstrate the effectiveness of \ours at mitigating extreme test-time distribution shifts in long-term forecasting scenarios, we verify \ours under the forecasting lengths of $H \in \{ 780, 840, 900\}$ on DLinear.
In Appendix, we plot the percentage of improvement in MSE achieved by \ours. 
Applying \ours exhibits a noticeable jump in performance improvement when compared to $H =336$; we note that on the ETTh2 dataset, the degree of improvement increases by more than $8\%$. 

\subsection{Compatibility with Methods Addressing Non-stationarity in Pre-training time}
One of the mainstream approaches to mitigating non-stationarity in \tsf is to employ normalization and denormalization modules~\cite{RevIN, Dish-TS, SAN}. 
However, they address non-stationarity only in the pre-training stage using training distributions, and thus, they may not generalize to consistently changing test distributions.
As \ours is pluggable to any source forecasters in test-time, this compatibility can further enhance the robustness of source forecasters pre-trained with these widely adopted normalization modules.
\tab~\ref{tab:norm} presents the results of applying \ours on source forecasters equipped with RevIN~\cite{RevIN}, Dish-TS~\cite{Dish-TS}, or SAN~\cite{SAN}.
Across all datasets and architectures, \ours further improves the forecasting capability of these advanced source forecasters.
The strength of \ours in long-range time series forecasting is again demonstrated here.
For iTransformer with $H = 720$, \ours reduces the MSE of RevIN and SAN by 8.90\% and 9.39\% on average. Likewise, for DLinear with $H = 720$, \ours improves RevIN and SAN by 4.45\% and 2.72\% on average.

Interestingly, the normalization approaches significantly increase the test MSE of the source forecaster in some experimental settings,~\textit{e.g.}, from 0.367 to 1.071 in FreTS + RevIN on ETTm1 with $H = 96$.
This observation highlights that addressing non-stationarity only in the pre-training phase can fail to generalize on the changing test distributions. 
In the above-mentioned setting, \ours improves the performance of FreTS + RevIN by \textbf{66.39\%}, indicating that it can overcome this limitation of pre-training-based approaches. 

\begin{table}[t]
    \footnotesize
    \renewcommand{\arraystretch}{0.9}
    \setlength{\tabcolsep}{4pt}
    \centering
    \caption{Test MSE with and without \ours on Chronos, a foundation model pre-trained on a massive corpus of time series data. We report test MSE for $H = 96$.}
    \vspace{-0.5em}
    \label{tab:foundation}
    \begin{tabular}{c|ccccc}
    \toprule
    Models & \ours& ETTh1 & ETTm1 & ETTh2 & ETTm2 \\
    \midrule
    \multirow{2}{*}{Chronos-small} & \xmark & 0.795 & 1.317 & 0.904 & 0.609 \\
     & \cmark & \textbf{0.624} & \textbf{0.858} & \textbf{0.611} & \textbf{0.492} \\
    \midrule
    \multirow{2}{*}{Chronos-base} & \xmark & 0.770 & 1.385 & 1.357 & 0.934 \\
     & \cmark & \textbf{0.611} & \textbf{0.761} & \textbf{0.708} & \textbf{0.668} \\
    \midrule
    \multirow{2}{*}{Chronos-large} & \xmark & 0.838 & 1.397 & 0.485 & 0.459 \\
     & \cmark & \textbf{0.635} & \textbf{0.772} & \textbf{0.477} & \textbf{0.431} \\
    \bottomrule
    \end{tabular}
\end{table}

\begin{table}[t]
    \footnotesize
    \renewcommand{\arraystretch}{0.9}
    \centering
    \caption{Comparison of test MSE with the state-of-the-art online \tsf methods: FSNet and OneNet for $H = 720$.}
    \vspace{-0.5em}
    \label{tab:online}
    \begin{tabular}{c|ccccc}
    \toprule
     & ETTh1 & ETTm1 & ETTh2 & ETTm2 & Exchange \\
    \midrule
    FSNet & 0.615 & 1.641 & 0.788 & 0.431 & 1.464 \\
    OneNet & \textbf{0.620} & 1.593 & 0.543 & 0.410 & 0.977 \\
    \midrule
    \ours & 0.640 & \textbf{0.512} & \textbf{0.356} & \textbf{0.289} & \textbf{0.704} \\
    \bottomrule
    \end{tabular}
    \vspace{-1em}
\end{table}

\subsection{Unleashing the Knowledge of Foundation Models}
Recently, \tsf foundation models pre-trained on up to billions of time steps~\cite{garza2023timegpt, TimesFM, ansari2024chronos, goswami2024moment} have shown promising forecasting performance.  
Here, we demonstrate that~\ours can further improve the performance of such a powerful source forecaster.
According to~\tab~\ref{tab:foundation},~\ours significantly improves the test MSE of Chronos~\cite{ansari2024chronos} on ETT datasets by up to 45\%.
The performance improvement is observed consistently on varying model sizes: small, base, and large.
We highlight that ETT datasets were not included in pre-training data, demonstrating that \ours effectively adapts \tsf foundation models to unseen time series data streams.
The compatibility of~\ours with foundation models alludes that it can adapt their predictions effectively without overwriting the rich semantic information encoded in these models.

\begin{figure}[t]
    \centering
    \includegraphics[width=\linewidth]{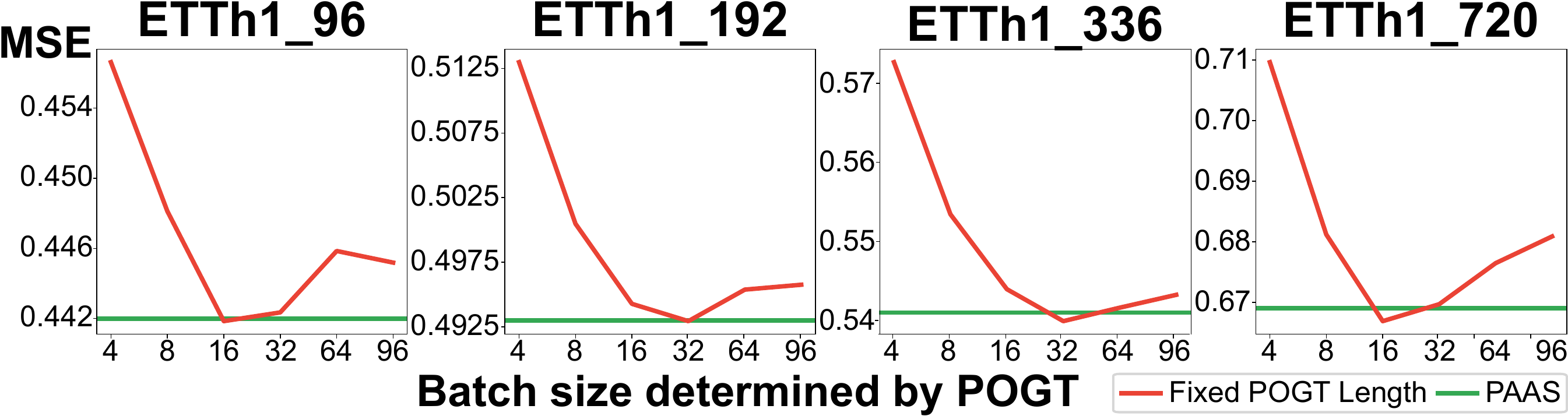}
    \vspace{-1.5em}
    \caption{Comparison of \paas against using a fixed POGT length on the ETTh1 dataset.}
    \label{fig:batchsize}
\end{figure}

\begin{table}[t]
    \footnotesize
    \renewcommand{\arraystretch}{0.9}
    \centering
    \caption{Forecasting errors as different modules are adapted in iTransformer. ``Norm.'' denotes the layer normalization.}
    \vspace{-0.5em}
    \label{tab:gcm}
    \begin{tabular}{c|cc|cc}
    \toprule
    \multirow{2}{*}{Modules to Adapt} & \multicolumn{2}{c|}{ETTh1} & \multicolumn{2}{c}{ETTh2} \\
     & MSE & MAE & MSE & MAE\\
    \midrule
    None (baseline) & 0.786 & 0.657 & 0.415 & 0.441 \\
    \midrule
    Norm. & 0.776 & 0.657 & 0.409 & 0.436 \\
    Encoder & 0.840 & 0.678 & 0.414 & 0.438 \\
    Decoder & 0.808 & 0.664 & 0.410 & 0.436 \\
    All & 0.832 & 0.673 & 0.414 & 0.438 \\
    \midrule
    GCM (\ours) & \textbf{0.704} & \textbf{0.627} & \textbf{0.393} & \textbf{0.425} \\
    \bottomrule
    \end{tabular}
\vspace{-1em}
\end{table}

\subsection{Comparison with Online \tsf Methods}
As mentioned in~\textbf{Related Works}, online \tsf is another line of research that leverages sequentially arriving data to update forecasters, but it differs from of~\tsftta in that online \tsf trains forecasters from scratch, whereas~\tsftta adapts a pre-trained source forecaster.
Here, we compare \ours with online~\tsf methods to corroborate that~\ours has tangible advantages over online \tsf.
\tab~\ref{tab:online} presents test MSE of the state-of-the-art online \tsf methods: FSNet~\cite{FastSlow} and OneNet~\cite{OneNet}, in the long-term forecasting scenario of $H = 720$. 
In most experimental settings, \ours significantly outperforms online \tsf methods.
The empirical superiority of~\ours to online~\tsf can be attributed to the following factors: 1) proactive adaptation of forecaster using periodicity-aware POGT, 2) the timely adjustment of predictions to reflect adjacent distribution shifts, and 3) preservation of the knowledge in the source forecaster through the use of auxiliary non-stationarity-aware GCM modules.

\subsection{Analysis of Each Technical Component in \ours}
We explore alternative design choices for two main technical components in \ours.
First, we study the effect of replacing \paas with a fixed POGT length.
Even though experiments are conducted with DLinear on all seven datasets, due to the page limit,~\fig~\ref{fig:batchsize} only shows the results on the ETTh1 dataset.
Extended results and the range of POGT lengths computed by \paas are in \suppl. 
\fig~\ref{fig:batchsize} shows that too short or long POGT curtails the effect of \ours, supporting the motivation behind dynamically adjusting its length.
Second, we demonstrate the effectiveness of introducing GCM.
\tablename~\ref{tab:gcm} presents forecasting errors for iTransformer with $H=720$ when other modules inside iTransformer are adapted. 
Modifying internal modules results in reduced performance improvement. In some cases, the forecasting performance drops below the baseline, likely due to the overwriting of core semantics in the source forecaster.
The effectiveness of each component in~\ours is shown through ablation studies in Appendix.
Lastly, to validate that \ours can be deployed at test-time without significant hyper-parameter tuning, we demonstrate its robustness to changes in hyper-parameters in Appendix. 

\section{Conclusion}
To address the ever-changing distributions in non-stationary time series, we propose \ours, a pioneering \tsftta framework that dynamically adapts the source forecaster at test-time while preserving the knowledge of the source forecaster. Using \paas, the partially-observed ground truth with semantically meaningful patterns is acquired for proactive adaptation. Then GCM, which considers both local and global temporal distribution shifts is adapted to address changing distributions. \ours serves as a dataset- and model-agnostic framework, demonstrated by thorough experimental results and analyses. The \tsftta framework pioneered in our work paves a new avenue toward sustainable deployment of state-of-the-art time series forecasters.

\section{Acknowledgments}
This work was supported by Institute of Information \& communications Technology Planning \& Evaluation (IITP) grant funded by the Korea government (MSIT) [No.RS-2021-II211343, Artificial Intelligence Graduate School Program (Seoul National University)],
the National Research Foundation of Korea (NRF) grant funded by the Korea government (MSIT)
(No. 2022R1A3B1077720, 2022R1A5A708390811), the BK21 FOUR program of the Education
and the Research Program for Future ICT Pioneers, Seoul National University in 2024, Hyundai Motor Company, and Samsung Electronics Co., Ltd (IO240124-08661-01).

\bibliography{aaai25}

\clearpage
\section*{Appendix} 
\setcounter{section}{0}
\renewcommand\thesection{A\arabic{section}}
\setcounter{table}{0}
\renewcommand{\thetable}{A\arabic{table}}
\setcounter{figure}{0}
\renewcommand{\thefigure}{A\arabic{figure}}
\setcounter{equation}{0}
\renewcommand{\theequation}{A\arabic{equation}}

\appendix

\section{Algorithm of \ours}
\algo~\ref{alg:code} summarizes the overall pipeline of \ours.

\section{Detailed Explanations on Datasets}
\label{suppl:details}
\tablename~\ref{tab:dataset} shows the characteristics of the seven widely used public \tsf benchmark datasets used throughout experiments. (Train, Validation, Test) shows the number of time steps in the train, validation, and test set, respectively. ADF shows the results of the Augmented Dickey-Fuller Test~\cite{elliott1992efficient}, which quantify the non-stationarity of each dataset. A higher ADF test statistic suggests greater non-stationarity, indicating prevalent distribution shifts.

\begin{table}[h]
\setlength{\tabcolsep}{4pt}
\centering
\caption{Characteristics of the seven widely used public \tsf benchmark datasets.}
\vspace{-1em}
\label{tab:dataset}

\begin{tabular}{l|cccc}
\toprule
\textbf{Dataset} & \textbf{Var.} & \textbf{Time Steps} & \textbf{Sampling Rate} & \textbf{ADF} \\ \midrule
ETTh1 & 7 & 17420 & 1 hour & -5.91 \\ 
ETTm1 & 7 & 69680 & 1 hour & -14.98 \\ 
ETTh2 & 7 & 17420 & 15 minutes & -4.13 \\
ETTm2 & 7 & 69680 & 15 minutes & -5.66 \\
Exchange & 8 & 7588 & 1 day & -1.90 \\
Illness & 7 & 966 & 1 week & -5.33 \\
Weather & 21 & 52696 & 10 minutes & -26.68 \\
\bottomrule
\end{tabular}
\vspace{-1em}
\end{table} 

\section{Training Details}
\label{suppl:train_details}
A hyper-parameter search was conducted for each source forecaster, exploring learning rates of 1e-3, 1e-4, and 1e-5, and weight decay values of 0.0, 1e-3, and 1e-4. For each combination, models were trained using three different seeds. The hyper-parameter combination that resulted in the lowest validation MSE loss was selected for use. 
Pre-training was performed for 30 epochs with a batch size of 64, ensuring that the validation MSE loss had sufficiently saturated. We employed the Adam optimizer~\cite{Adam} and adjusted the learning rate using cosine learning rate scheduling~\cite{loshchilov2016sgdr}. All experiments were conducted using a single NVIDIA A40 GPU. We referred the configuration (e.g., number of layers) of each source forecaster to the Time Series Library~\cite{Timesnet}.
When incorporating the normalization-based approaches (RevIN~\cite{RevIN}, Dish-TS~\cite{Dish-TS}, and SAN~\cite{SAN}) for pre-training the source forecaster, the hyper-parameter searches for RevIN and Dish-TS were conducted in the same manner as for training the baseline source forecaster. For SAN, a two-stage pre-training process is required, involving the training of the statistics prediction module. According to the settings in the original paper, the statistics prediction module was trained for 10 epochs, with a hyper-parameter search for learning rates of 1e-3 and 1e-4. 
When performing TSF-TTA with TAFAS, the learning rate was searched among 5e-3, 3e-3, 1e-3, 5e-4, and 1e-4. The initial value of the gating parameter $\alpha$ was searched among 0.01, 0.05, 0.1, and 0.3. For reproducibility, we will make our code publicly available in the final version.

\begin{algorithm}[ht]
\caption{PyTorch-style Pseudocode for \ours}
\label{alg:code}
\definecolor{codeblue}{rgb}{0.25,0.5,0.5}
\definecolor{codekw}{rgb}{0.85, 0.18, 0.50}
\lstset{
  backgroundcolor=\color{white},
  basicstyle=\fontsize{9pt}{9pt}\ttfamily\selectfont,
  columns=fullflexible,
  breaklines=true,
  captionpos=b,
  commentstyle=\fontsize{7.5pt}{7.5pt}\color{codeblue},
  keywordstyle=\fontsize{7.5pt}{7.5pt}\color{codekw},
}
\begin{lstlisting}[language=python]
# L: look-back window length
# forecaster: source forecaster
# in_GCM: input GCM
# out_GCM: output GCM

forecaster.requires_grad_(False)
do_PAAS = True
test_batch = []

for test_input in test_loader:
    # PAAS
    if do_PAAS:
        period = PAAS(test_input)
        bsz = 0
        do_PAAS = False
        
    # get mini-batch based on PAAS
    if bsz < period + 1:
        test_batch.append(test_input)
        bsz += 1
        if bsz == period + 1:
            do_PAAS = True
        else:
            continue
            
    # get POGT
    test_batch = torch.stack(test_batch)
    POGT = test_batch[-1][-:period]
    
    # adapt GCM
    input_cali = in_GCM(test_input)
    pred = forecaster(input_cali)
    pred_cali = out_GCM(pred)
    l_p = MSE(pred_cali[:period], POGT)
    if full_gt_available:
        l_f = MSE(full_pred, full_GT)
    else:
        l_f = 0.0
    l_tafas = l_p + l_f
    l_tafas.backward()
    optimizer.zero_grad()
    optimizer.step()
    
    # PA
    with torch.no_grad():
        pred_adapted = out_CGM(forecaster(in_GCM(test_batch)))
        for i in range(bsz - 1):
            pred_cali[i, period-i:] = pred_adapted[i, period-i:]
            
    test_batch = []


def PAAS(test_input):
    # test_input: (L, C)
    test_input -= test_input.mean(dim=0)
    amplitude = abs(fft(test_input))
    var_idx = (amplitude ** 2).sum(dim=0).argmax()
    freq = amplitude[:, var_idx].argmax()
    p = test_input.shape[0] // freq
    
    return p

\end{lstlisting}
\end{algorithm}
    
\clearpage

\section{Additional Related Works}
\label{suppl:related}
\subsection{Time Series Forecasting Models}
As time series forecasting has become a pivotal application in various industries, diverse time series forecasting architectures have been developed to accurately predict future time steps~\cite{wen2022transformers, masini2023machine}. They range from traditional statistical methods~\cite{arima, GaussianProcess, hyndman2018forecasting} to deep neural network-based architectures, including Transformers~\cite{Transformer, Informer, LogSparseTransformer, Pyraformer, iTransformer, Autoformer, PatchTST, Crossformer, Reformer, Fedformer, wu2022flowformer}, Linear layers~\cite{DLinear, RLinear, OLS}, and MLPs~\cite{FreTS, MICN, TSMixer, TiDE, FITS, wang2024timemixer, Nhits}. 
Transformer-based models aim to capture temporal dependencies as well as inter-variable dependencies utilizing the attention mechanism. To address the quadratic time and memory complexity of self-attention operation, a line of work has been proposed to modify the self-attention module to be more efficient, facilitating long-term forecasting~\cite{Informer, Reformer, LogSparseTransformer}. Some works have revised the Transformer architecture to better exploit the properties of time series, such as sub-series periodicity or frequency information~\cite{Autoformer, Fedformer}. Other works remain the architecture untouched, but consider how to input time series by patching or inverting~\cite{PatchTST, iTransformer}. On the other hand, in response to a recent work that raised questions to the modeling capability of Transformer-based \tsf models~\cite{DLinear}, a series of Linear~\cite{DLinear, RLinear, OLS} and MLP-based architectures~\cite{FreTS, MICN, TSMixer, TiDE, FITS, wang2024timemixer, Nhits} have been developed, achieving comparable or outperforming \tsf capabilities compared to Transformer-based models. Still, there is no consensus on the most representative \tsf architecture, making the model-agnosticism of the \tsftta framework more desirable. 
More recently, time series foundation models, pre-trained on up to billions of time steps, have shown promising forecasting capabilities~\cite{ansari2024chronos, TimesFM, goswami2024moment, garza2023timegpt}. The emergence of foundation models highlighted exploiting rich semantic information encoded in pre-trained models on unseen data streams.

\subsection{Test-Time Adaptation (\tta)}
The goal of test-time adaptation is to adapt a pre-trained source model to distributionally shifted inputs encountered during testing, thereby improving generalization performance on unseen test distributions. Test-time adaptation has primarily evolved in classification tasks, leveraging the characteristic components of classification tasks~\cite{TENT, EATA, SAR, DEYO, NOTE}.
\tta methods commonly rely on the entropy of class probabilities, which is only applicable to classification tasks and infeasible to \tsf, which is a regression task. 
TENT~\cite{TENT} minimizes the entropy of the target class probability by updating the channel-wise affine transformation of the normalization layer. EATA~\cite{EATA} builds on the entropy minimization framework by proposing a sample filtering method for sample-efficient entropy minimization. SAR~\cite{SAR} considers additional scenarios in classification tasks, such as label imbalance, and proposes sharpness-aware entropy minimization. DEYO~\cite{DEYO} combines entropy minimization with image-specific data augmentation, utilizing pseudo label probability for the transformed test input. 
Moreover, they do not consider the properties of time series data, such as instance-wise distribution shifts within a window input and the global temporal dependency across streaming window data.
Although NOTE~\cite{NOTE} considers the temporal correlation between streaming images, it still relies on an entropy-based approach and does not account for the distribution shifts that can occur within a single instance window in time series data.
Additionally, they are not fully model-agnostic because they adapt specific types of normalization layers (e.g., Batch Norm~\cite{BatchNorm} or Layer Norm~\cite{LayerNorm}) and are inapplicable to models without normalization layers.
Given that \tsf has been developed based on various architectural advances, model dependence significantly limits the generality of \tsftta framework. 
\clearpage

\section{Additional Experimental Results}
\begin{figure}[t]
    \centering
    \includegraphics[width=\linewidth]{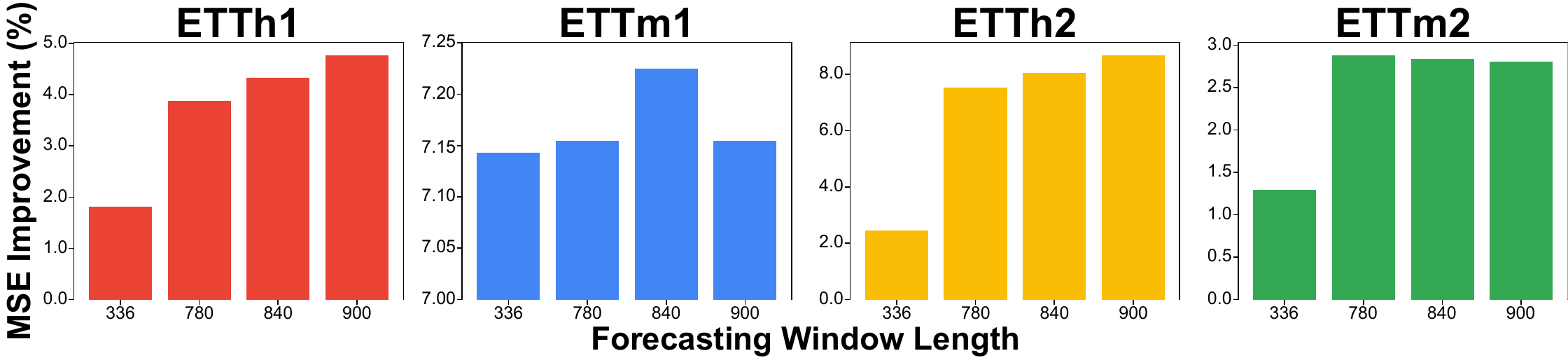}
    \caption{Improvements of MSE (\%) as the forecasting window length increases.}
    \label{fig:mse_improvement}
\end{figure}

\fig~\ref{fig:mse_improvement} presents the effectiveness of \ours in long-term forecasting scenarios that we stated in the Experiments Section in the manuscript.
\label{suppl:exp}

\begin{table}[t]
    \footnotesize
    \renewcommand{\arraystretch}{1.0}
    \centering
    \caption{Comparison of average test MSE between \ours and baseline models. Both \ours and the baselines are adapted using the same input with sequentially arriving test data. Baselines include DLinear and DLinear pre-trained with Dish-TS.}
    \vspace{-0.5em}
    \label{tab:baseline_1}
    \begin{tabular}{c|ccccc}
    \toprule
     & ETTh1 & ETTm1 & ETTh2 & ETTm2 & Exchange \\
    \midrule
    DLinear & 0.553 & 0.476 & 0.313 & 0.222 & 0.350 \\
    +\ours & 0.538 & 0.450 & 0.304 & 0.219 & 0.337 \\
    \midrule
    DLinear & \multirow{2}{*}{0.550} & \multirow{2}{*}{0.452} & \multirow{2}{*}{0.308} & \multirow{2}{*}{0.234} & \multirow{2}{*}{0.300} \\
    w/ Dish-TS & & & & & \\
    +\ours & 0.525 & 0.446 & 0.291 & 0.221 & 0.256 \\
    \bottomrule
    \end{tabular}
\end{table}

\begin{table}[t]
    \footnotesize
    \renewcommand{\arraystretch}{1.0}
    \centering
    \caption{Forecasting errors for different ($\mu$, $\alpha$).}
    \label{tab:robustness}
    \begin{tabular}{c|cc}
    \toprule
    ($\mu$, $\alpha$) & MSE & MAE \\
    \midrule
    (1e-3, 0.01) & 0.669 \scriptsize $\pm$ 0.002 & 0.600 \scriptsize $\pm$ 0.002\\
    (1e-3, 0.05) & 0.669 \scriptsize $\pm$ 0.001 & 0.598 \scriptsize $\pm$ 0.000\\
    (1e-3, 0.1) & 0.691 \scriptsize $\pm$ 0.001 & 0.609 \scriptsize $\pm$ 0.001\\
    \midrule
    (5e-4, 0.01) & 0.682 \scriptsize $\pm$ 0.000 & 0.602 \scriptsize $\pm$ 0.000\\
    (5e-4, 0.05) & 0.671 \scriptsize $\pm$ 0.001 & 0.599 \scriptsize $\pm$ 0.000\\
    (5e-4, 0.1) & 0.670 \scriptsize $\pm$ 0.001 & 0.599 \scriptsize $\pm$ 0.000\\
    \midrule
    (1e-4, 0.05) & 0.683 \scriptsize $\pm$ 0.001 & 0.601 \scriptsize $\pm$ 0.001\\
    (1e-4, 0.1) & 0.676 \scriptsize $\pm$ 0.001 & 0.599 \scriptsize $\pm$ 0.000\\
    (1e-4, 0.3) & 0.669 \scriptsize $\pm$ 0.001 & 0.599 \scriptsize $\pm$ 0.000\\
    \bottomrule
    \end{tabular}
\end{table}

\begin{table}[t]
    \footnotesize
    \renewcommand{\arraystretch}{1.0}
    \setlength{\tabcolsep}{3pt}
    \centering
    \caption{Ablation study on each component of \ours. The ``-'' denotes the excluded component from \ours.}
    \label{tab:ablation}
    \begin{tabular}{cl|cc|cc|cc}
    \toprule
    \multicolumn{1}{c}{} & & \multicolumn{2}{c|}{Exchange} & \multicolumn{2}{c|}{ETTh1} & \multicolumn{2}{c}{ETTh2} \\
    \multicolumn{1}{c}{} & & MSE & MAE& MSE & MAE& MSE & MAE \\
    \midrule    
    \multicolumn{2}{c|}{Baseline} & 0.844 & 0.692 & 0.786 & 0.662 & 0.415 & 0.441 \\
    \midrule
    \multirow{3}{*}{TAFAS} & - \paas & 0.795 & 0.672 & 0.740 & 0.645 & 0.393 & 0.425 \\
    & - GCM & 0.979 & 0.756 & 1.320 & 0.839 & 0.491 & 0.480 \\
    & - PA & 0.809 & 0.679 & 0.706 & 0.628 & 0.392 & 0.425 \\
    \midrule
    \multicolumn{2}{c|}{\ours} & \textbf{0.773} & \textbf{0.665} & \textbf{0.704} & \textbf{0.627} & \textbf{0.393} & \textbf{0.425} \\
    \bottomrule
    \end{tabular}
\end{table}

\begin{table}[t]
    \centering
    \caption{Range of POGT length $p$ constructed via \paas.}
    \label{tab:appen_batch_range}
    \begin{tabular}{l|c}
    \toprule
    Dataset & Range of $p$ \\
    \midrule
    ETTh1 & 12-96 \\
    ETTm1 & 32-96 \\
    ETTh2 & 8-96 \\
    ETTm2 & 6-96 \\
    Exchange & 48-96 \\
    Illness & 36-36 \\
    Weather & 2-96 \\
    \bottomrule
    \end{tabular}
\end{table}

\begin{sidewaystable*}[ht]
    \tiny
    \setlength{\tabcolsep}{5.5pt}
    \centering
    \caption{Test MAE and standard deviations on multivariate time series forecasting datasets with and without \ours across various \tsf architectures: Transformer-based (iTransformer~\cite{iTransformer}, PatchTST~\cite{PatchTST}), Linear-based (DLinear~\cite{DLinear}, OLS~\cite{OLS}), and MLP-based (FreTS~\cite{FreTS}, MICN~\cite{MICN}). $+$\ours denotes whether the \ours framework is applied to the corresponding source forecaster. 
    }
    \resizebox{\textwidth}{!}{
    \begin{tabular}{cc|cccc|cccc|cccc}
        \toprule
        \multicolumn{2}{c}{} & \multicolumn{4}{c}{Transformer-based} & \multicolumn{4}{c}{Linear-based} & \multicolumn{4}{c}{MLP-based} \\
        \cmidrule(lr){3-6} \cmidrule(lr){7-10} \cmidrule(lr){11-14}
        \multicolumn{2}{c}{Models} & \multicolumn{2}{c}{iTransformer} & \multicolumn{2}{c}{PatchTST} & \multicolumn{2}{c}{DLinear} & \multicolumn{2}{c}{OLS} & \multicolumn{2}{c}{FreTS} & \multicolumn{2}{c}{MICN} \\
        \multicolumn{2}{c}{$+$ \ours} & \xmark & \cmark & \xmark & \cmark & \xmark & \cmark & \xmark & \cmark & \xmark & \cmark & \xmark & \cmark \\
        \bottomrule
        \toprule
        \multirow{4}{*}{\rotatebox[origin=c]{90}{ETTh1}} & 96 & 0.448 (0.000) & 0.443 (0.000) & 0.450 (0.001) & 0.444 (0.000) & 0.446 (0.000) & 0.442 (0.000) & 0.446 (0.000) & 0.444 (0.000) & 0.447 (0.000) & 0.444 (0.000) & 0.459 (0.000) & 0.453 (0.000) \\
         & 192 & 0.494 (0.000) & 0.489 (0.000) & 0.489 (0.001) & 0.483 (0.000) & 0.483 (0.000) & 0.480 (0.000) & 0.483(0.000) & 0.482 (0.000) & 0.486 (0.000) & 0.483 (0.000) & 0.503 (0.004) & 0.497 (0.002) \\
         & 336 & 0.532 (0.001) & 0.532 (0.000) & 0.520 (0.001) & 0.519 (0.001) & 0.515 (0.000) & 0.516 (0.000) & 0.514 (0.000) & 0.512 (0.000) & 0.527 (0.000) & 0.525 (0.000) & 0.546 (0.002) & 0.544 (0.001) \\
         & 720 & 0.662 (0.006) & 0.627 (0.000) & 0.621 (0.006) & 0.621 (0.005) & 0.605 (0.000) & 0.598 (0.002) & 0.605 (0.000) & 0.599 (0.000) & 0.619 (0.000) & 0.610 (0.000) & 0.642 (0.006) & 0.633 (0.005) \\
        \midrule
        \multirow{4}{*}{\rotatebox[origin=c]{90}{ETTm1}} & 96 & 0.405 (0.003) & 0.387 (0.001) & 0.403 (0.002) & 0.397 (0.001) & 0.393 (0.000) & 0.381 (0.000) & 0.393 (0.000) & 0.382 (0.000) & 0.391 (0.000) & 0.386 (0.001) & 0.412 (0.009) & 0.397 (0.004) \\
         & 192 & 0.438 (0.001) & 0.430 (0.000) & 0.435 (0.000) & 0.428 (0.000) & 0.428 (0.000) & 0.416 (0.000) & 0.429 (0.000) & 0.416 (0.000) & 0.426 (0.001) & 0.423 (0.001) & 0.431 (0.000) & 0.423 (0.000) \\
         & 336 & 0.475 (0.002) & 0.466 (0.001) & 0.468 (0.000) & 0.461 (0.000) & 0.467 (0.000) & 0.452 (0.000) & 0.467 (0.000) & 0.452 (0.000) & 0.463 (0.001) & 0.458 (0.000) & 0.471 (0.001) & 0.461 (0.001) \\
         & 720 & 0.523 (0.001) & 0.511 (0.001) & 0.517 (0.001) & 0.509 (0.000) & 0.515 (0.000) & 0.497 (0.000) & 0.515 (0.000) & 0.499 (0.000) & 0.510 (0.001) & 0.504 (0.001) & 0.532 (0.002) & 0.516 (0.001) \\
        \midrule
        \multirow{4}{*}{\rotatebox[origin=c]{90}{ETTh2}} & 96 & 0.330 (0.001) & 0.329 (0.001) & 0.321 (0.001) & 0.320 (0.001) & 0.315 (0.000) & 0.314 (0.000) & 0.317 (0.000) & 0.316 (0.000) & 0.322 (0.000) & 0.321 (0.001) & 0.321 (0.001) & 0.320 (0.001) \\
         & 192 & 0.365 (0.001) & 0.362 (0.001) & 0.357 (0.002) & 0.353 (0.002) & 0.352 (0.000) & 0.350 (0.000) & 0.352 (0.000) & 0.349 (0.000) & 0.359 (0.000) & 0.355 (0.000) & 0.357 (0.001) & 0.354 (0.001) \\
         & 336 & 0.392 (0.001) & 0.386 (0.001) & 0.386 (0.002) & 0.382 (0.001) & 0.379 (0.000) & 0.374 (0.000) & 0.379 (0.000) & 0.372 (0.000) & 0.386 (0.000) & 0.378 (0.000) & 0.386 (0.004) & 0.378 (0.001) \\
         & 720 & 0.441 (0.001) & 0.425 (0.001) & 0.438 (0.004) & 0.427 (0.003) & 0.433 (0.000) & 0.417 (0.000) & 0.434 (0.000) & 0.412 (0.000) & 0.441 (0.000) & 0.419 (0.000) & 0.434 (0.003) & 0.421 (0.003) \\
        \midrule
        \multirow{4}{*}{\rotatebox[origin=c]{90}{ETTm2}} & 96 & 0.265 (0.000) & 0.263 (0.001) & 0.262 (0.000) & 0.262 (0.000) & 0.264 (0.000) & 0.262 (0.000) & 0.264 (0.000) & 0.263 (0.000) & 0.260 (0.001) & 0.259 (0.000) & 0.265 (0.000) & 0.264 (0.000) \\
         & 192 & 0.296 (0.000) & 0.292 (0.000) & 0.295 (0.000) & 0.294 (0.000) & 0.290 (0.000) & 0.289 (0.000) & 0.291 (0.000) & 0.290 (0.000) & 0.291 (0.000) & 0.290 (0.000) & 0.293 (0.000) & 0.291 (0.000) \\
         & 336 & 0.331 (0.001) & 0.324 (0.000) & 0.325 (0.001) & 0.323 (0.000) & 0.319 (0.000) & 0.317 (0.000) & 0.320 (0.000) & 0.317 (0.000) & 0.321 (0.000) & 0.319 (0.000) & 0.322 (0.000) & 0.320 (0.000) \\
         & 720 & 0.373 (0.001) & 0.366 (0.000) & 0.371 (0.001) & 0.367 (0.001) & 0.365 (0.000) & 0.359 (0.000) & 0.365 (0.000) & 0.359 (0.000) & 0.366 (0.001) & 0.359 (0.001) & 0.366 (0.000) & 0.361 (0.000) \\
         \midrule
        \multirow{4}{*}{\rotatebox[origin=c]{90}{Exchange}} & 96 & 0.207 (0.000) & 0.208 (0.000) & 0.202 (0.001) & 0.200 (0.001) & 0.197 (0.000) & 0.197 (0.000) & 0.197 (0.000) & 0.195 (0.000) & 0.201 (0.000) & 0.198 (0.000) & 0.200 (0.000) & 0.198 (0.000) \\
         & 192 & 0.298 (0.000) & 0.293 (0.000) & 0.301 (0.001) & 0.296 (0.001) & 0.292 (0.000) & 0.289 (0.000) & 0.293 (0.000) & 0.289 (0.000) & 0.297 (0.001) & 0.291 (0.001) & 0.303 (0.001) & 0.297 (0.000) \\
         & 336 & 0.415 (0.001) & 0.389 (0.003) & 0.431 (0.010) & 0.395 (0.005) & 0.408 (0.000) & 0.387 (0.001) & 0.409 (0.000) & 0.384 (0.000) & 0.412 (0.000) & 0.393 (0.002) & 0.420 (0.004) & 0.403 (0.005) \\
         & 720 & 0.692 (0.001) & 0.665 (0.000) & 0.691 (0.004) & 0.690 (0.004) & 0.688 (0.001) & 0.684 (0.000) & 0.687 (0.000) & 0.577 (0.000) & 0.689 (0.004) & 0.668 (0.018) & 0.839 (0.015) & 0.726 (0.031) \\
        \midrule
        \multirow{4}{*}{\rotatebox[origin=c]{90}{Illness}} & 24     & 0.906 (0.008) & 0.907 (0.008) & 0.841 (0.003) & 0.841 (0.003)       & 1.055 (0.002) & 1.042 (0.002) & 1.008 (0.001) & 0.993 (0.000)       & 0.946 (0.002) & 0.946 (0.002) & 1.029 (0.002) & 1.030 (0.002)\\
         & 36 & 0.922 (0.006) & 0.922 (0.006) & 0.860 (0.007) & 0.860 (0.007) & 1.026 (0.001) & 1.001 (0.005) & 1.000 (0.001) & 0.948 (0.000) & 0.961 (0.007) & 0.961 (0.007) & 1.062 (0.002) & 1.063 (0.001) \\
         & 48 & 0.939 (0.001) & 0.933 (0.002) & 0.854 (0.018) & 0.854 (0.018) & 1.030 (0.000) & 0.015 (0.000) & 1.005 (0.002) & 0.957 (0.000) & 0.947 (0.004) & 0.913 (0.004) & 0.923 (0.004) & 0.913 (0.005) \\
         & 60 & 0.895 (0.007) & 0.872 (0.008) & 0.862 (0.005) & 0.862 (0.005) & 1.046 (0.001) & 1.033 (0.001) & 1.029 (0.000) & 0.975 (0.000) & 0.934 (0.004) & 0.915 (0.004) & 0.914 (0.005) & 0.906 (0.003) \\
        \midrule
        \multirow{4}{*}{\rotatebox[origin=c]{90}{Weather}} & 96 & 0.220 (0.002) & 0.219 (0.002) & 0.215 (0.001) & 0.219 (0.001) & 0.235 (0.000) & 0.235 (0.000) & 0.234 (0.000) & 0.237 (0.000) & 0.220 (0.000) & 0.220 (0.000) & 0.219 (0.001) & 0.219 (0.000) \\
         & 192 & 0.260 (0.000) & 0.257 (0.001) & 0.257 (0.000) & 0.260 (0.000) & 0.270 (0.000) & 0.267 (0.000) & 0.270 (0.000) & 0.273 (0.000) & 0.259 (0.000) & 0.261 (0.001) & 0.262 (0.001) & 0.264 (0.001) \\
         & 336 & 0.301 (0.000) & 0.300 (0.000) & 0.297 (0.000) & 0.300 (0.000) & 0.306 (0.000) & 0.303 (0.000) & 0.306 (0.000) & 0.305 (0.000) & 0.298 (0.000) & 0.295 (0.000) & 0.302 (0.001) & 0.303 (0.001) \\
         & 720 & 0.352 (0.001) & 0.352 (0.001) & 0.346 (0.000) & 0.356 (0.000) & 0.353 (0.000) & 0.348 (0.000) & 0.353 (0.000) & 0.351 (0.000) & 0.347 (0.000) & 0.344 (0.000) & 0.346 (0.001) & 0.356 (0.003) \\
        \bottomrule
        \label{tab:appen_main_mae}
    \end{tabular}
    }
\end{sidewaystable*}

\begin{figure*}[t]
    \centering
    \includegraphics[width=\linewidth]{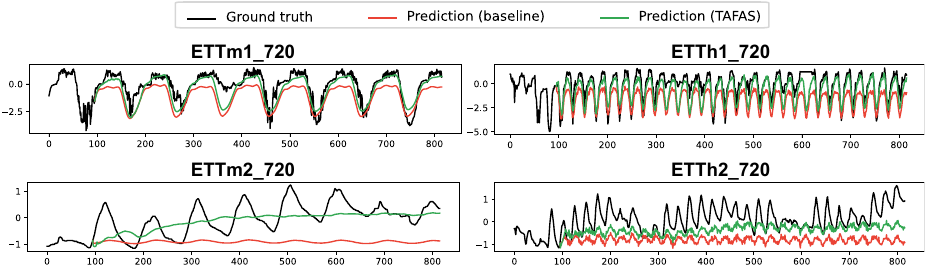}
    \caption{Visualization of forecasting results with and without \ours. The top row illustrates that TAFAS effectively adapts to both low-frequency (top left) and high-frequency (top right) dominant patterns within the look-back window. The bottom row highlights the promising aspect of TAFAS in significantly more challenging scenarios characterized by pronounced global distribution shifts.}
    \label{fig:qualitative}
\end{figure*}

\subsection{Comparison with Baselines using POGT}
To further verify the effectiveness of \ours, we have conducted experiments where the baselines are also adapted using the same input as \ours. The baselines includee DLinear and DLinear with Dish-TS. All results are reported in terms of MSE averaged over $H \in \{ 96, 192, 336, 720 \}$. According to \tab~\ref{tab:baseline_1}, the baselines fall short of TAFAS even when they use POGT for further training. This demonstrates that \ours consistently outperforms the baselines by effectively leveraging sequentially arriving test data.

\subsection{Hyperparameter Robustness Analysis}
\label{appen_hyperparam}
To validate that \ours can be deployed at test-time without significant hyper-parameter tuning, we show its robustness to changes in hyper-parameters.~\tab~\ref{tab:robustness} presents test MSE and MAE for different combinations of the two hyper-parameters involved in \ours: the test-time learning rate $\mu$ and initialization value for gating parameter $\alpha$. The experiments were conducted using DLinear on the ETTh1 dataset with $H = 720$. \ours constantly achieves low MSE and MAE across different parameter settings with consistently low standard deviations, indicating that it is reasonably robust against changes in hyper-parameters.
\subsection{Component-wise Ablation Studies}
\label{appen_ablation}
\begin{figure*}[t]
    \centering
    \includegraphics[width=\linewidth]{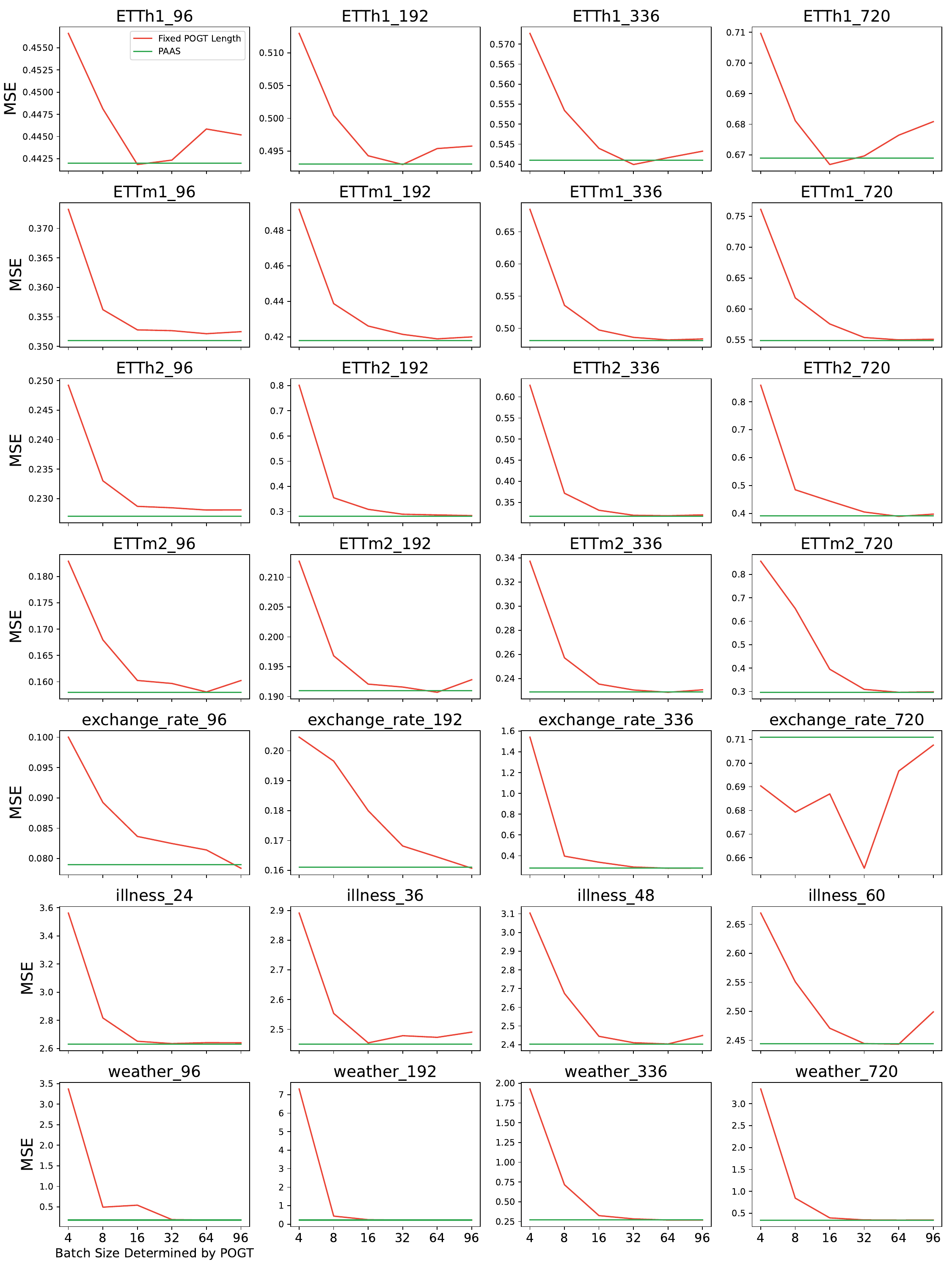}
    \caption{Comparison of test MSE when using \paas against fixed length POGT.}
    \label{fig:appen_batchsize}
\end{figure*}

\tab~\ref{tab:ablation} presents ablative experiments aimed at demonstrating the necessity of various components of \ours for successfully performing \tsftta. When GCM is excluded, the entire source forecaster is adapted to maintain model-agnosticity of \tsftta. In the absence of \paas, a test batch size of 64 is arbitrarily selected. The experiments are conducted for iTransformer with $H = 720$. Each row in~\tab~\ref{tab:ablation} indicates which component was removed for ablation; besides the removed component, the rest of \ours remains unchanged. The results show that removing each component of \ours subsequently degrades the performance of \ours, validating its effectiveness. \fig~\ref{fig:appen_batchsize} illustrates the MSE values for each dataset when test batches are adaptively composed using \paas compared to using fixed batch sizes. In most instances, \paas consistently achieves superior performance. The variation in effective batch sizes across different datasets underscores the importance of identifying an optimal batch size for each dataset. \paas demonstrates its capability to configure these batch sizes at test time effectively. \tablename~\ref{tab:appen_batch_range} shows the range of test batch sizes derived through \paas for each dataset. 

The results show that introducing each component of \ours subsequently reduces the MSE, validating its effectiveness. \\

\subsection{Qualitative Analysis}
\fig~\ref{fig:qualitative} shows the qualitative comparison of forecasting results with and without \ours for iTransformer. 
In the top row, the baseline produces repetitive patterns that resemble the latest periodic pattern within the look-back window.
In contrast, \ours preemptively adapts the source forecaster to changing distributions and outputs significantly more accurate predictions. 
Furthermore, TAFAS effectively adapts to both low-frequency (top left) and high-frequency (top right) dominant patterns.
The two panels on the bottom row highlight the particular advantages of \ours in more challenging long-term forecasting scenarios where test data digress noticeably from historical data, resulting in a more extreme distribution shift. 
In the examples in these two panels, the time series in the prediction window visible differs from that in the look-back window, which is indicative of a global shift in distribution.
Here, the baseline merely repeats the pattern within the look-back window, while~\ours effectively captures the global distribution shift.

\subsection{MAE and Standard Deviations of \ours}
\tablename~\ref{tab:appen_main_mae} presents the MAE and standard deviations for the three different runs. The standard deviations were rounded to the fourth decimal place. When \ours is applied, most standard deviations are 0.000, demonstrating the stability of \ours in test-time adaptation. 

\clearpage

\end{document}